\documentclass[10pt,twocolumn,letterpaper]{article}

\usepackage{wacv}              % To produce the CAMERA-READY version
\usepackage{graphicx}
\usepackage{amsmath}
\usepackage{amssymb}
\usepackage{booktabs}

\usepackage[pagebackref,breaklinks,colorlinks]{hyperref}

\usepackage{multirow}
\usepackage{array}
\usepackage{makecell}
\usepackage{graphbox}
\usepackage[normalem]{ulem}

\usepackage{pifont}

\newcolumntype{P}[1]{>{\centering\arraybackslash}p{#1}}
\newcolumntype{M}[1]{>{\centering\arraybackslash}m{#1}}
\newcommand{\cmark}{\ding{51}}
\newcommand{\xmark}{\ding{55}}

\usepackage{algorithm}
\usepackage{algorithmic}
\usepackage{setspace}

\usepackage[capitalize]{cleveref}
\crefname{section}{Sec.}{Secs.}
\Crefname{section}{Section}{Sections}
\Crefname{table}{Table}{Tables}
\crefname{table}{Tab.}{Tabs.}

\def\wacvPaperID{1202} % *** Enter the WACV Paper ID here
\def\confName{WACV}
\def\confYear{2025}

\begin{document}

\title{SPACE: SPAtial-aware Consistency rEgularization for anomaly detection in Industrial applications}

\author{
    Daehwan Kim\textsuperscript{$1*$}\qquad
    Hyungmin Kim\textsuperscript{$1*$}\qquad
    Daun Jeong\textsuperscript{$1$}\qquad
    Sungho Suh\textsuperscript{$2,3$}\qquad
    Hansang Cho\textsuperscript{$1$}\\
	$^1$Samsung Electro-Mechanics, Suwon, Republic of Korea\\
	$^2$German Research Center for Artificial Intelligence (DFKI), Kaiserslautern, Germany\\
        $^3$Department of Computer Science, RPTU Kaiserslautern-Landau, Kaiserslautern, Germany\\
        {\tt\small \{daehwan85.kim, du33.jeong, hansang.cho\}@samsung.com}, {\tt\small sungho.suh@dfki.de}
}
\maketitle

%%%%%%%%%
\def\thefootnote{*}\footnotetext{Equal contribution.}

%%%%%%%%% ABSTRACT
\begin{abstract}
In this paper, we propose SPACE, a novel anomaly detection methodology that integrates a Feature Encoder (FE) into the structure of the Student-Teacher
method. The proposed method has two key elements: Spatial Consistency regularization Loss (SCL) and Feature converter Module (FM). SCL prevents overfitting in student models by avoiding excessive imitation of the teacher model. Simultaneously, it facilitates the expansion of normal data features by steering clear of abnormal areas generated through data augmentation. This dual functionality ensures a robust boundary between normal and abnormal data. The FM prevents the learning of ambiguous information from the FE. This protects the learned features and enables more effective detection of structural and logical anomalies. Through these elements, SPACE is available to minimize the influence of the FE while integrating various data augmentations.
In this study, we evaluated the proposed method on the MVTec LOCO, MVTec AD, and VisA datasets. Experimental results, through qualitative evaluation, demonstrate the superiority of detection and efficiency of each module compared to state-of-the-art methods.
\end{abstract}

%%%%%%%%% BODY TEXT
\vspace{-0.5cm}
\section{Introduction}\label{sec:intro}
Anomaly detection (AD) is a crucial task, particularly challenging in domains where acquiring abnormal 
samples for the training
set is impractical, such as medical~\cite{zhou2021proxy, zhao2021anomaly, wolleb2022diffusion, wei2018anomaly} and industrial applications~\cite{zheng2022focus, carrera2016defect, tao2022deep, napoletano2018anomaly}. 
The task is conventionally categorized into two groups: reconstruction-based \cite{baur2019deep, tang2020anomaly, davletshina2020unsupervised, nguyen2019anomaly, bergmann2022beyond, gong2019memorizing, perera2019ocgan} and embedding similarity-based methods~\cite{defard2021padim, roth2022towards, hyun2023reconpatch}.
The reconstruction-based methods regenerate input data and measure the reconstruction error for anomaly detection while embedding similarity-based approaches focus on the distance or similarity between embedded features to identify outliers.
% ---------------------------------------------------------------
\begin{figure}[t]
    \centering
    \resizebox{1.0\columnwidth}{!}{
    \setlength{\tabcolsep}{1pt}
    \begin{tabular}{cc}
        \includegraphics[width=\columnwidth]{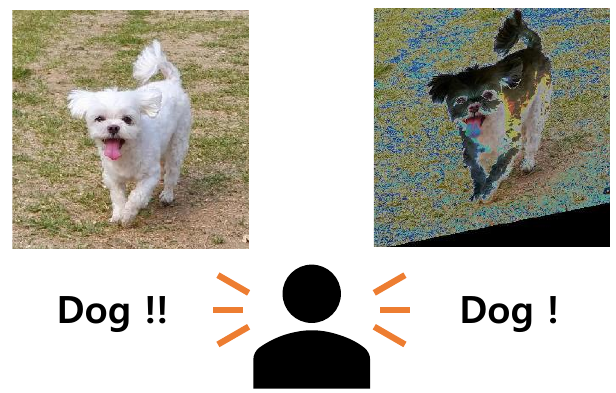} &\includegraphics[width=\columnwidth]{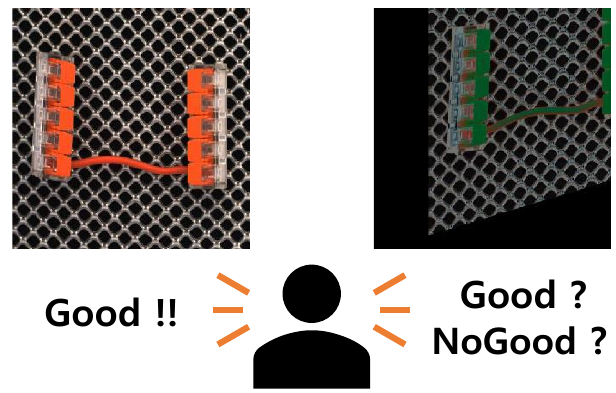}\\
        \fontsize{0.65cm}{0.65cm}\selectfont{(a) Classification} &\fontsize{0.65cm}{0.65cm}\selectfont{(b)Anomaly detection}\\
    \end{tabular}}
    \vspace{-0.25cm}
    \caption{The ambiguities of applying strong augmentation for industrial data.} 
    \vspace{-0.5cm}
    \label{fig:method_scheme}    
\end{figure}
% ---------------------------------------------------------------
\begin{figure*}[t]
    \centering
    \resizebox{0.95\linewidth}{!}{
    \setlength{\tabcolsep}{1pt}
    \begin{tabular}{ccc}
        \includegraphics[width=0.25\linewidth]{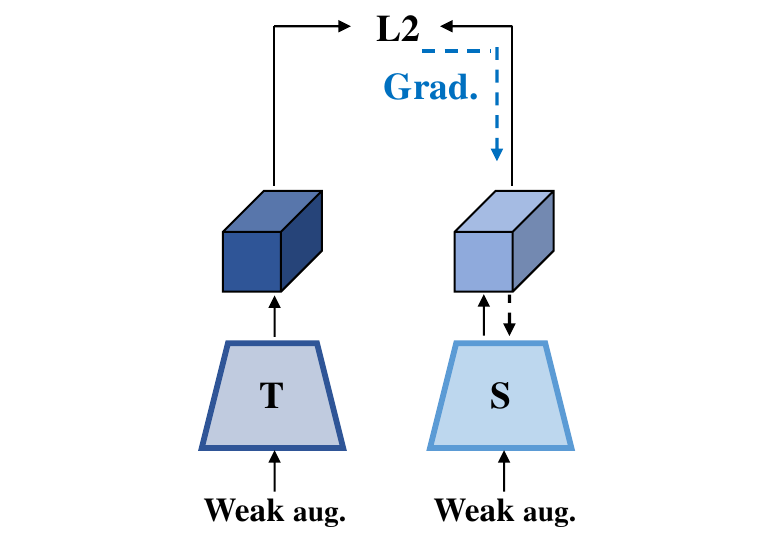}
        &\includegraphics[width=0.25\linewidth]{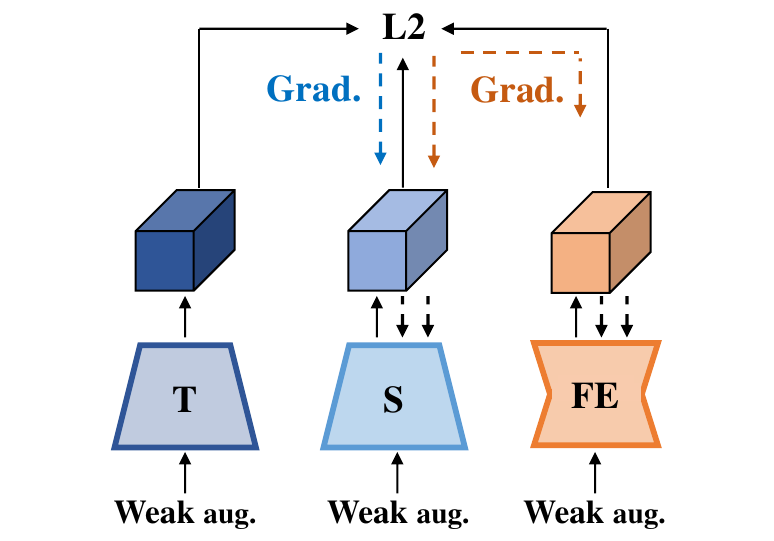}
        &\includegraphics[width=0.25\linewidth]{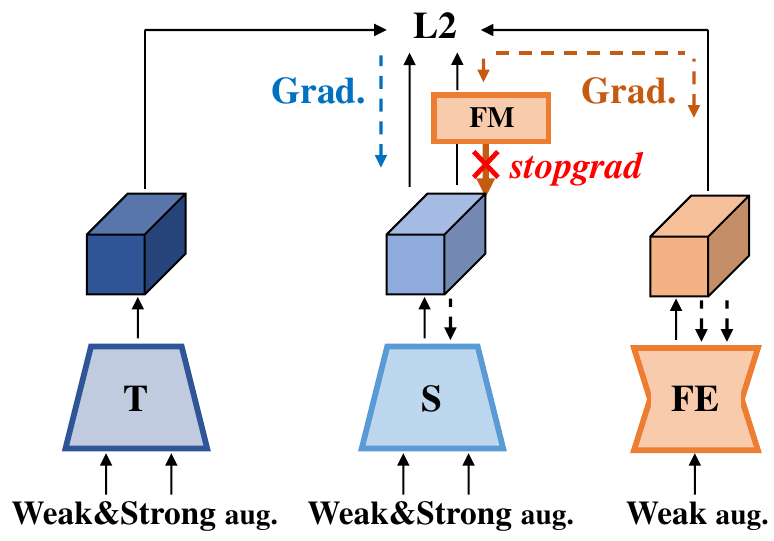}\\
        \fontsize{0.225cm}{0.225cm}\selectfont{(a) S-T} 
        &\fontsize{0.225cm}{0.225cm}\selectfont{(b) S-T with FE} 
        &\fontsize{0.225cm}{0.225cm}\selectfont{(c) SPACE}
    \end{tabular}}
    \vspace{-0.3cm}
    \caption{\textbf{Comparison of existing and proposed approaches for anomaly detection:} (a) represents the basic S-T structure, (b) shows the structure combining S-T and FE, and (c) depicts the SPACE structure proposed.} 
    \label{fig:method_scheme_arch}
    \vspace{-0.5cm}
\end{figure*}

Recently, Student-Teacher (S-T) approaches have demonstrated remarkable performance in 
AD
~\cite{bergmann2020uninformed, yamada2022reconstructed, zhang2023destseg, xiao2021unsupervised}.
This method uses an ImageNet pre-trained model as a teacher and trains the student exclusively on normal data to detect anomalies. However, collecting sufficient data for effective student model training is challenging.

Data augmentation methods, widely adopted in tasks like classification and segmentation \cite{moreno2020improving, mikolajczyk2018data, nalepa2019data, olsson2021classmix, zoph2020learning, kaur2021data}, face limitations in industrial
AD datasets~\cite{bergmann2021mvtec, zou2022spot}.
The main challenge is that abnormal regions are small portions of the entire image, while the rest is normal. Applying strong augmentations to normal data makes it more similar to anomalies rather than normals. \cref{fig:method_scheme}(a) illustrates why applying strong augmentation is particularly challenging in industrial data. There are several approaches to leveraging limited augmentations for this task.
RegAD~\cite{huang2022registration} uses only weak augmentations on normal samples for the task. Also, CutPaste~\cite{li2021cutpaste} and DRAEM~\cite{zavrtanik2021draem} employ augmentations to create abnormals, rather than normal data, to detect anomalies.

The recently released dataset, MVTec LOCO~\cite{bergmann2022beyond} requires detection not only of structural abnormalities in small areas but also of logical anomalies.
These logical anomalies refer to cases where individual objects may be normal, but the overall image level is abnormal.
For example, in the pushpin dataset, a situation where there is more than one pin or no pin at a location that should have one is defined as a logical abnormality, even if the pins have normal shapes.
Identifying logical anomalies with existing approaches can be challenging.
To address the issue, EfficientAD~\cite{batzner2023efficientad} and GCAD~\cite{bergmann2022beyond} train the student model to mimic the features of the Feature Encoder (FE). However, this may still introduce unclear features, potentially leading to false positives.

To enhance both structural and logical performance by leveraging both weak and strong augmentations, we propose a framework called Spatial-aware Consistency Regularization for Anomaly (SPACE). The framework has two key elements, which are Spatial Consistency regularization Loss (SCL) and Feature converter Module (FM).
SCL serves two crucial roles. One is that prevents the student model from overly mimicking the teacher model during the S-T learning process, thus mitigating overfitting. The other is that helps the student model broaden the boundaries of normal
features by avoiding anomalous regions generated through
augmentations
and selectively learning features.
FM prevents ambiguous information conveyed by the FE from being learned by the student model. This helps protect the learned features and enables more effective detection of both structural and logical anomalies. \cref{fig:method_scheme_arch}
shows the differences between
SPACE
and the existing structures.
We conduct extensive evaluations of the proposed method on the MVTec AD,
MVTec LOCO, and VisA datasets. Experimental results demonstrate its superiority compared to the state-of-the-art methods in both detection and the effectiveness of each module with qualitative evaluations.

Our contributions in this paper are fourfold:
% ---------------------------------------------------------------
\begin{itemize}
    \item We introduce a novel loss function, SCL, which enables the utilization of both strong and weak augmentations for
    the AD task.
    \vspace{-0.25cm}
     \item We propose a simple yet powerful FM that improves not only logical but also structural
     AD performance.
     \vspace{-0.25cm}
     \item SPACE outperforms the state-of-the-art methods in terms of Image-level AUROC on the MVTec AD, MVTec LOCO, and VisA datasets.
     \vspace{-0.25cm}
     \item We provide extensive ablation studies to validate the effectiveness of SCL and FM.
\end{itemize}
% ---------------------------------------------------------------
\begin{figure*}[ht]
    \centering
    \resizebox{0.8\linewidth}{!}{
    \setlength{\tabcolsep}{1pt}
    \begin{tabular}{c}
        \includegraphics[width=1.\linewidth, trim=0cm 0cm 0cm 1cm]{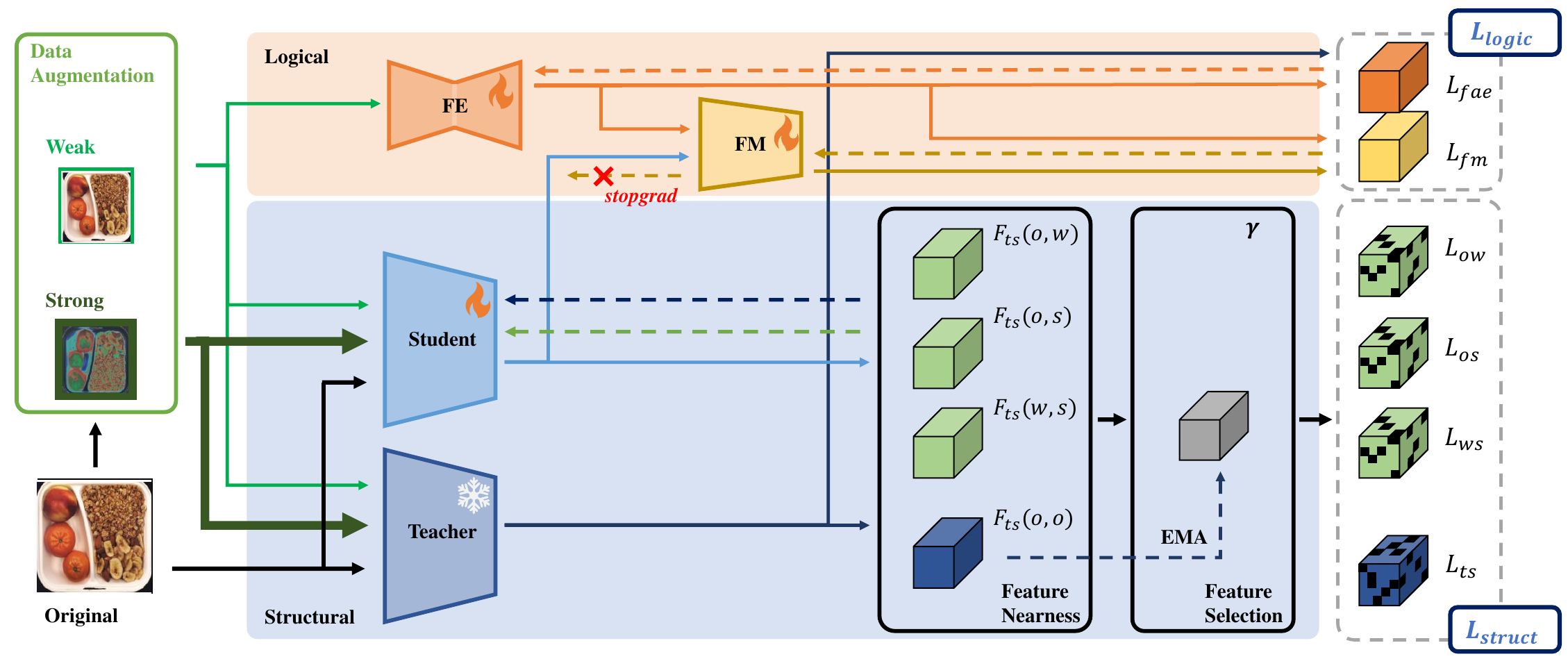}\\
    \end{tabular}}
    \vspace{-0.5cm}
    \caption{\textbf{The overall architecture:} The method combines a structural branch, detecting fine-grained anomalies through consistency regularization, and a logical branch, focusing on shape anomalies using the FM.}
    \vspace{-0.5cm}
    \label{fig:method_scenarios}
\end{figure*}

% ---------------------------------------------------------------
\vspace{-0.25cm}
\section{Related Works}\label{sec:realatedworks}
\subsection{Anomaly Detection (AD)}\label{sec:ad}
Anomaly detection methods aim to predict abnormal characteristics by relying on the information contained within normal data. These methods can be categorized into two types: embedding-based and reconstruction-based. The embedding-based method utilizes embedding vectors obtained from a pre-trained model on generic datasets like ImageNet or fine-tunes the embeddings for specific domains.
SPADE~\cite{cohen2020sub} and PaDiM~\cite{defard2021padim} utilized hierarchical feature extraction and memory banks to detect anomalies based on differences from the distribution of normal embedding vectors.
PatchCore~\cite{roth2022towards} simplified the detection process and enhanced efficiency by utilizing subsampling of the core set.
On the other hand, in reconstruction-based anomaly detection, a prominent method involves the use of autoencoders~\cite{gong2019memorizing, chow2020anomaly, said2020network, an2015variational, chen2018autoencoder, kim2022spatial}. These approaches aim to identify the differences between the input and the reconstructed image, particularly in cases where abnormalities cannot be effectively reconstructed using the learned normal patterns.
Another reconstruction method involves generative adversarial networks~\cite{jiang2019gan, zenati2018efficient, xia2022gan}, which train one model to generate data similar to a typical dataset and another model to distinguish between real and generated data.
Recently, diffusion-based
methods~\cite{zhang2023unsupervised, wolleb2022diffusion, munoz2021unsupervised} have been employed to
enhance image quality.
The approaches restored more complex and higher-quality images, thereby enhancing AD performance. The recent advent of the MVTec LOCO~\cite{bergmann2022beyond} has expanded the AD problem to include both structural and logical anomalies. 
The most recent methods, GCAD~\cite{bergmann2022beyond} and EfficientAD~\cite{batzner2023efficientad}, achieved outstanding performance by combining the S-T
network
and FE.
Our framework
also drew inspiration from this structure.
% ---------------------------------------------------------------
\subsection{Consistency Regularization}\label{sec:ssl}
Consistency regularization was first introduced in the field of semi-supervised learning through the $\Pi$-model~\cite{sajjadi2016regularization}, and it enhances the robustness of models by maintaining stable outputs across various input transformations. The regularization has undergone further refinement and expansion by incorporating diverse augmentation techniques. MixMatch~\cite{berthelot2019mixmatch} presented a new direction in this field by combining Mixup~\cite{zhang2017mixup}
with consistency regularization. Notably, FixMatch~\cite{sohn2020fixmatch} achieved remarkable results by applying consistency regularization with combinations of both weak and strong augmentations, as well as pseudo-labeling~\cite{pseudo2013simple, grandvalet2004semi}.
Recently, 
ConMatch~\cite{kim2022conmatch} has further enhanced the efficiency of
the
regularization by incorporating two types of strong
augmentations.
We extend the use of regularization, employed in semi-supervised learning, to AD by incorporating spatial recognition functionality.

% ---------------------------------------------------------------
\section{Method}\label{sec:method}
Our method aims to learn normal representation spaces by considering adjustments in feature distances, assuming that abnormal features are fixed. \cref{fig:method_scenarios} describes the overall structure of our framework, which consists of two components. The first component focuses on spatial-aware consistency regularization and is designed to detect structural anomalies, including object and texture abnormalities, by using SCL between the model's representations of weakly and strongly augmented data. Data augmentation can enable the model to generalize effectively to diverse anomaly scenarios for robust anomaly detection. 
The second module is designed to detect logical abnormalities by reconstructing normal data using the FE and the FM.
Detecting logical abnormalities is useful for comprehensive anomaly detection, as it provides insights into overarching patterns or irregularities that may not be discernible at the local level.
\subsection{Spatial-aware Consistency Loss (SCL)}\label{sec:sadl}
To detect a structural abnormality, the fixed pre-trained teacher model $g_t$ and the learned student model $g_s$ are compared.
The goal of our structure is for $g_s$ to learn the essential features of $g_t$ while ensuring that normal feature pairs are close together.
In this context, we exploit two additional views, the strongly augmented $x_s$ and the weakly augmented inputs $x_w$, in conjunction with the original input data $x_{o}$ to achieve our objective. These perspectives serve as positive features.
The method employs original-to-weak, original-to-strong, and weak-to-strong consistency loss to learn meaningful representations of normal data. And these are denoted as $\tilde{D}(x_o,x_w)$, $\tilde{D}(x_o,x_s)$, and $D(x_w,x_s)$, respectively. $\tilde{D}(x_o,x_w)$ is defined as follows:
%-------------------------------------------------------------------------
\begin{equation}
\begin{aligned}
    \tilde{D}(x_o,x_w) = ({\scriptstyle stopgrad}(g_s(x_{o})) - g_s(x_{w}))_{c,w,h}^{2}, 
\end{aligned}
\label{eq:d_w}
\end{equation}
where ${\scriptstyle stopgrad}(g_s(x_{o}))$ means $g_s(x_{o})$ is treated as a constant vector in this term. $(g_s(x_{w}))_{c,w,h}^2$ denotes the element-wise squaring of $g_s(x_{w})$. $\tilde{D}(x_o,x_s)$ is represented similarly to \cref{eq:d_w}. 
$c$, $w$, and $h$ indicate the channel, width, and height of the feature, respectively.
And $\tilde{D}(x_w,x_s)$ is defined as follows:
%-------------------------------------------------------------------------
\begin{equation}
\begin{aligned}
    D(x_w,x_s) = (g_s(x_{w}) - g_s(x_{s}))_{c,w,h}^{2}, 
\end{aligned}
\label{eq:d_ws}
\end{equation}

Employing a naive combination of losses in this way is well-known to be helpful for specific tasks, such as classification. However, as shown in~\cref{fig:method_scheme}, utilizing the combination undermines the anomaly detection task due to ambiguities in the strongly augmented data. To address the issue, the model needs to update feature distances selectively, concentrating on features deemed important. In this manner, we introduce a criterion $\Upsilon$ to identify necessary features that rely on distances between $g_t(x_{o})$ and $g_s(x_{o})$. Namely, $\Upsilon$ aims to discern whether a given feature falls within the range of confidently recognized normal features.
%-------------------------------------------------------------------------
\begin{equation}
\begin{aligned}
    \Upsilon = ema((g_t(x_{o}) - g_s(x_{o}))_{c,w,h}^{2}),
\end{aligned}
\label{eq:upsilon}
\end{equation}
%-------------------------------------------------------------------------
\noindent Here, $ema$ indicates the exponential moving average (EMA) for the feature differences.
In other words, $\Upsilon$ plays a similar role to the score threshold in semi-supervised learning, and we calculate a normal feature nearness $F_{ts}$, which is similar to confidence scores.
%-------------------------------------------------------------------------
\begin{equation}
\begin{aligned}
    F_{ts}(x_o,x_w)= (g_t(x_{o}) - g_s(x_{w}))_{c,w,h}^{2}, 
\end{aligned}
\label{eq:f_ts_w}
\end{equation}

In this manner, the original-to-weak consistency loss $\mathcal{L}_{ow}$ and the original-to-strong consistency loss $\mathcal{L}_{os}$ leverage features in the vicinity of the criterion and are described as follows, respectively:
%-------------------------------------------------------------------------
\begin{equation}
\begin{aligned}
   \mathcal{L}_{ow}=&~\mathbb{E}[\mathbb{I}(F_{ts}(x_o,x_w)<\Upsilon) \odot \tilde{D}(x_o,x_w)],
\end{aligned}
\label{eq:l_w}
\end{equation}
%-------------------------------------------------------------------------
\begin{equation}
\begin{aligned}
   \mathcal{L}_{os}=&~\mathbb{E}[\mathbb{I}(F_{ts}(x_o,x_s)<\Upsilon) \odot \tilde{D}(x_o,x_s)],
\end{aligned}
\label{eq:l_s}
\end{equation}
%-------------------------------------------------------------------------
\noindent where $\odot$ denotes the element-wise multiplication and $\mathbb{I}(\cdot)$ means a feature size mask that satisfies element-wise inequality.

And 
the weak-to-strong consistency loss $\mathcal{L}_{ws}$ is designed using two masks.
%-------------------------------------------------------------------------
\begin{equation}
    \begin{aligned}
        \mathcal{L}_{ws}=&~\mathbb{E}[\mathbb{I}(F_{ts}(x_o,x_w)<\Upsilon)\\
        &~\odot\mathbb{I}(F_{ts}(x_o,x_s)<\Upsilon)~\odot~D(x_w,x_s)],
    \end{aligned}
\label{eq:l_mw}
\end{equation}

Additionally, SCL has adopted the feature distillation loss $\mathcal{L}_{ts}$ to improve generalization and prevent overfitting.
In this process, using
the teacher,
normal features are distilled to
the student 
by minimizing the element-wise squared distance, and this is defined as follows:
%-------------------------------------------------------------------------
\begin{equation}
\begin{aligned}
    F_{ts}(x_o, x_o)= (g_t(x_{o}) - g_s(x_{o}))_{c,w,h}^{2}, 
\end{aligned}
\label{eq:f_ts}
\end{equation}
%-------------------------------------------------------------------------
\begin{equation}
\begin{aligned}
    \mathcal{L}_{ts}= \mathbb{E}[\mathbb{I}(F_{ts}(x_o,x_o)>\Upsilon) \odot F_{ts}(x_o,x_o)],  
\end{aligned}
\label{eq:l_ts}
\end{equation}
%-------------------------------------------------------------------------
\noindent 
where $\Upsilon$ also serves to focus on essential features and prevent unnecessary learning during training.

To sum up, SCL aims to detect structural abnormalities and is defined as follows:
%-------------------------------------------------------------------------
\begin{equation}
\begin{aligned}
    \mathcal{L}_{structural}= \mathcal{L}_{scl} = \mathcal{L}_{ts} + \lambda_1(\mathcal{L}_{ow}+\mathcal{L}_{os}+\mathcal{L}_{ws}),
\end{aligned}
\label{eq:l_obj}
\end{equation}
where $\lambda_1$ denotes a hyper-parameter for balancing losses.

\noindent\textbf{Reason for using both Weak and Strong Augmentations:}
We are selectively training features based on EMA. Nevertheless, when using strong augmentation, there is a possibility that features outside the normal range may also be updated. To prevent this situation, we use weak augmentation. Since weak augmentation is almost similar to normal data, more features are updated compared to strong augmentation, thus accelerating the learning process. This method helps to quickly expand the range of normal features and prevents abnormal features caused by strong augmentation from being included in the learning process.
%-------------------------------------------------------------------------
\subsection{Feature converter Module (FM)}\label{sec:dstill_module}
Instead of comparing the characteristics of FE $f_{ae}$ and $g_{s}$ to detect logical anomalies, we compare the characteristics of $f_{ae}$ and FM $f_{fm}$.
$f_{ae}$ learns $g_{t}$ with the following equation.
%-------------------------------------------------------------------------
\begin{equation}
\begin{aligned}
    \mathcal{L}_{fae}=\mathbb{E}[(f_{ae}(x_w) - g_t(x_w))_{c,w,h}^2],
\end{aligned}
\label{eq:l_ae}
\end{equation}

\noindent $f_{ae}$ has a bottleneck structure, which means it does not perfectly depict $g_{t}$, but learns the overall features. This can create differences with $g_{s}$ and increase false positives. To address this issue, $f_{fm}$ is introduced to transform the sharp features of $g_{s}$ into similar overall features as $f_{ae}$, and its equation is as follows:
%-------------------------------------------------------------------------
\begin{equation}
\begin{aligned}
    Z_{fm}(x_w)=(f_{ae}(x_w) - f_{fm}({\scriptstyle stopgrad}(g_s(x_w))))_{c,w,h}^2,
\end{aligned}
\label{eq:z_fm}
\end{equation}

\noindent Here, before $f_{fm}$,
${\scriptstyle stopgrad}(\cdot)$
prevents the unclear feature information from $f_{ae}$ from flowing into $g_{s}$. The loss for detecting logical abnormalities is as follows:
%-------------------------------------------------------------------------
\begin{equation}
\begin{aligned}
    \mathcal{L}_{fm}=\mathbb{E}[\mathbb{I}(Z_{fm}(x_w)\geq d_{hard}) \odot Z_{fm}(x_w)],\\
\end{aligned}
\label{eq:l_fm}
\end{equation}
%-------------------------------------------------------------------------
\begin{equation}
\begin{aligned}
    \mathcal{L}_{logical}=\mathcal{L}_{fae} +\lambda_2\mathcal{L}_{fm},\\
\end{aligned}
\label{eq:l_log}
\end{equation}
where $d_{hard}$ is the $q^{th}$ quantile of the input tensor $Z_{fm}(x_w)$. This means that instead of utilizing the entire set of features for learning, only the values with significant differences are targeted for the learning process. $\lambda_2$ means a hyper-parameter for balancing losses.

In conclusion, the total loss, incorporating both structural and logical abnormality loss, is defined as:
%-------------------------------------------------------------------------
\begin{equation}
\begin{aligned}
    \mathcal{L}_{total}=\mathcal{L}_{structural} +\mathcal{L}_{logical}.\\
\end{aligned}
\label{eq:l_total}
\end{equation}
%-------------------------------------------------------------------------
\begin{table*}[ht]
\footnotesize
\begin{center}
\setlength{\tabcolsep}{1pt}
\renewcommand{\arraystretch}{1.1}
\begin{tabular}
{p{0.11\linewidth}P{0.11\linewidth}P{0.12\linewidth}P{0.14\linewidth}P{0.11\linewidth}P{0.11\linewidth}P{0.14\linewidth}P{0.11\linewidth}}
    \toprule
    Category &S-T\cite{bergmann2020uninformed} &SPADE\cite{cohen2020sub} &PatchCore\cite{roth2022towards} &GCAD\cite{bergmann2022beyond} &SLSG\cite{yang2023slsg} &EfficientAD\cite{batzner2023efficientad} &SPACE\\
    \midrule
    \midrule
    Breakfast &(\underline{68.6},49.6) &(78.2,\underline{37.2}) &(81.3,46.0) &(83.9,50.2)
    &(\textbf{88.9},\textbf{65.9}) &(87.5,60.4) &(87.5,65.5)\\
    Bottle &(91.0,81.1) &(\underline{88.3},80.4) &(95.6,\underline{71.0}) &(99.4,91.0) &(99.1,82.0) &(99.5,93.4) &(\textbf{100},\textbf{93.8})\\
    Pushpins &(74.9,52.3) &(\underline{59.3},\underline{23.4}) &(72.3,44.7) &(86.2,73.9) &(95.5,\textbf{74.4}) &(96.4,62.3) &(\textbf{98.8},66.0)\\
    Screw bag &(71.2,60.2) &(\underline{53.2},\underline{33.1}) &(64.9,52.2) &(63.2,55.8) &(\textbf{79.4},47.2) &(74.3,\textbf{64.4}) &(78.3,62.4)\\
    Connector &(81.1,69.8) &(\underline{65.4},\underline{51.6}) &(82.4,58.6) &(83.9,\textbf{79.8}) &(88.5,66.9) &(95.3,73.5) &(\textbf{98.2},65.5)\\
    \midrule
    Mean &(77.3,62.6) &(\underline{68.8},\underline{45.1}) &(79.3,54.5) &(83.3,\textbf{70.1}) &(90.3,67.3) &(90.6,69.4) &(\textbf{92.6},69.7)\\ 
    \bottomrule
\end{tabular}
\end{center}
\vspace{-0.5cm}
\caption{The performance of various methods in anomaly detection and localization on MVTec LOCO dataset, using the Image-level AUROC(\%) and Pixel-level sPRO(\%) metrics, respectively. We computed the area under the sPRO curve up to a false positive rate of 0.05. The highest value is represented in bold, and the lowest value is underlined.}
\vspace{-0.25cm}
\label{tab:experiment_01}
\end{table*}
%-------------------------------------------------------------------------
\begin{table*}[t]
\begin{center}
\footnotesize
\setlength{\tabcolsep}{1pt}
\renewcommand{\arraystretch}{1.1}
\begin{tabular}
{p{0.11\linewidth}P{0.11\linewidth}P{0.12\linewidth}P{0.14\linewidth}P{0.11\linewidth}P{0.11\linewidth}P{0.14\linewidth}P{0.11\linewidth}}
    \toprule
    Category &S-T\cite{bergmann2020uninformed} &SPADE\cite{cohen2020sub} &PatchCore\cite{roth2022towards} &GCAD\cite{bergmann2022beyond} &SLSG\cite{yang2023slsg} &{EfficientAD}\cite{batzner2023efficientad} &SPACE\\
    \midrule
    \midrule
    Carpet &(95.3,97.7) &(-,97.5) &(98.7,\textbf{99.0}) &(63.6,95.6) &(99.0,96.0) &(99.4,96.1) &(\textbf{100},98.5)\\
    Grid &(98.1,99.2) &(-,93.7) &(98.2,98.7) &(91.1,98.0) &(\textbf{100},98.5) &(99.8,99.3) &(99.7,\textbf{99.4})\\
    Leather &(94.7,99.3) &(-,97.6) &(\textbf{100},99.3) &(72.0,98.4) &(\textbf{100},99.5) &(\textbf{100},\textbf{99.6}) &(\textbf{100},\textbf{99.6})\\
    Tile &(99.9,98.3) &(-,87.4) &(98.7,95.6) &(88.2,96.8) &(\textbf{100},\textbf{98.6}) &(\textbf{100},98.1) &(\textbf{100},98.5)\\
    Wood &(99.1,96.6) &(-,85.5) &(99.2,95.0) &(90.3,94.6) &(\textbf{99.6},96.8) &(99.4,96.6) &(99.2,\textbf{97.6})\\
    \midrule
    Bottle &(99.0,98.3) &(-,98.4) &(\textbf{100},98.6) &(98.4,96.8) &(99.4,\textbf{99.1}) &(\textbf{100},98.9) &(\textbf{100},99.0)\\
    Cable &(78.7,96.0) &(-,97.2) &(\textbf{99.5},\textbf{98.4}) &(92.5,97.4) &(98.3,97.4) &(97.9,98.8) &(96.6,98.2)\\
    Capsule &(92.5,98.6) &(-,99.0) &(98.1,98.8) &(77.3,97.5) &(95.5,95.9) &(98.8,99.1) &(\textbf{99.9},\textbf{99.6})\\
    Hazelnut &(99.1,98.8) &(-,\textbf{99.1}) &(\textbf{100},98.7) &(97.1,99.0) &(99.5,97.8) &(99.4,97.6) &(99.0,98.6)\\
    Metal nut &(89.1,\textbf{99.4}) &(-,98.1) &(\textbf{100},98.4) &(95.4,98.2) &(\textbf{100},98.9) &(98.9,98.3) &(98.6,97.8)\\
    Pill &(92.2,98.7) &(-,96.5) &(96.6,97.4) &(86.1,97.8) &(\textbf{99.2},98.0) &(94.5,\textbf{99.6}) &(99.0,98.7)\\
    Screw &(86.0,98.8) &(-,98.9) &(98.1,99.4) &(92.6,99.0) &(89.1,97.3 &(94.2,98.7) &(\textbf{98.5},\textbf{99.7})\\
    Toothbrush &(100,99.1) &(-,97.9) &(\textbf{100},98.7) &(\textbf{100},98.7) &(\textbf{100},\textbf{99.4}) &(99.9,95.0) &(\textbf{100},99.0)\\
    Transistor &(79.4,79.6) &(-,94.1) &(\textbf{100},96.3) &(99.8,97.9) &(97.3,92.5) &(99.9,95.0) &(\textbf{100},\textbf{98.0})\\
    Zipper &(94.4,98.9) &(-,96.5) &(99.4,98.8) &(93.0,96.1) &(\textbf{100},97.1) &(98.6,98.9) &(98.0,\textbf{99.1})\\    
    \midrule
    Mean &(93.2,97.1) &(85.5,96.0) &(99.1,98.1) &(89.1,97.4) &(98.5,97.5) &(98.7,98.2) &(\textbf{99.2},\textbf{98.8})\\ 
    \bottomrule
\end{tabular}
\end{center}
\vspace{-0.5cm}
\caption{The performance in anomaly detection and localization on the MVTec AD dataset, using the Image-level AUROC(\%) and Pixel-level AUROC(\%) metrics, respectively. The highest value is represented in bold.}
\vspace{-0.5cm}
\label{tab:experiment_02}
\end{table*}
%-------------------------------------------------------------------------
\vspace{-0.75cm}
\subsection{Anomaly Score}\label{sec:anomlay_score}
The proposed method computes both the structural and logical anomaly maps for the input $x$.
The differences between $g_t$ and $g_s$ create the local anomaly map $M_{structural}$, while $f_{ae}$ and $f_{fm}$ are used to calculate the global anomaly map $M_{logical}$.
The 
total anomaly map $M_{total}$ is composed by combining these two anomaly maps, and the formula is defined as follows:
%-------------------------------------------------------------------------
\begin{equation}
\begin{aligned}
    M_{structural}(x)= \frac{1}{C} \sum_{c=1}^{C}(g_t(x) - g_s(x))_{c,w,h}^{2}, 
\end{aligned}
\label{eq:m_structural}
\end{equation}
%-------------------------------------------------------------------------
\begin{equation}
\begin{aligned}
    M_{logical}(x)=\frac{1}{C} \sum_{c=1}^{C}(f_{ae}(x) - f_{fm}(g_s(x))_{c,w,h}^2,
\end{aligned}
\label{eq:m_logical}
\end{equation}
%-------------------------------------------------------------------------
\begin{equation}
\begin{aligned}
    M_{total}(x)=\frac{1}{2}\times M_{structural}(x) + \frac{1}{2}\times M_{logical}(x).
\end{aligned}
\label{eq:m_total}
\end{equation}
%-------------------------------------------------------------------------
\noindent where $g_t(x)$ is normalized using the mean and variance of the training data.
% The normalization methods for 
$M_{structural}(\cdot)$ and $M_{logical}(\cdot)$ are the same as in EfficientAD~\cite{batzner2023efficientad}, where the top 99.5\% and 90\% values of the anomaly scores in the validation data are used as the maximum and minimum values, respectively, for normalization. Ultimately, the anomaly score is determined using the maximum value of $M_{total}(x)$.
%-------------------------------------------------------------------------
\section{Experiments}\label{sec:experiments}
\subsection{Experimental Settings}\label{sec:settings}
\noindent\textbf{Comparison Methods:}
We compared the performance of SPACE with state-of-the-art algorithms including EfficientAD~\cite{batzner2023efficientad}, S-T~\cite{bergmann2020uninformed}, SPADE~\cite{cohen2020sub}, PatchCore~\cite{roth2022towards}, GCAD~\cite{bergmann2022beyond}, and SLSG~\cite{yang2023slsg}. SPADE and PatchCore utilize embedding-based methods where a network pre-trained on ImageNet is used to embed normal data for modeling. Then, anomaly detection is performed by comparing abnormal data with the modeled normal data. This approach directly utilizes features from ImageNet without
additional training. S-T employs a teacher-student framework where a network pre-trained on ImageNet serves as the teacher model, and only normal data from the data intended
for AD
is used to train the student.
AD
is then carried out by leveraging differences in features between the teacher and student networks when presented with abnormal data. GCAD and EfficientAD also utilize ImageNet pre-training and employ architectures consisting of two branches responsible for logical and structural anomalies. SLSG pre-trains its network using a masked autoencoder approach and trains it with simulated abnormal images to perform tasks such as segmentation.

\noindent\textbf{Datasets:}
We evaluate our method across 32 AD scenarios in
MVTec AD~\cite{bergmann2021mvtec}, MVTec LOCO~\cite{bergmann2022beyond}, and VisA~\cite{zou2022spot} datasets. 
MVTec AD is primarily used to test anomaly detection and localization in manufacturing fields. The dataset consists of five textures and ten object categories. VisA is divided into 12 subsets, each corresponding to a single object class. Recently, MVTec LOCO has been introduced to present more diverse and challenging problems to existing MVTec AD.
The dataset contains five categories, with a new type, logical anomaly, introduced.

\noindent\textbf{Evaluation metrics:}
For the evaluation, we measure
AD
performance across entire images using AUROC. To precisely measure the abnormality of each pixel, we compute the AUROC based on the ground truth mask. For the MVTec LOCO, pixel-level performance is assessed using the Saturated Per-Region Overlap (sPRO) metric proposed by GCAD to clarify logical anomalies. All results reported are of the weights of the last iteration.
\subsection{Implementation Details}\label{sec:details} 
Our proposed method aims to achieve high-speed inference.
For this reason, SPACE is based on EfficientAD-M.
We have adjusted the dimension of the last convolutional layer in the FE to 256 instead of 64 for improved representativeness. Furthermore, 
The FM
consists of three convolution layers with input and output channels of 384, using $3\times 3$ kernels and a stride of 1. The ReLU activation function is applied, but no activation function is used in the last layer. Additionally, a skip connection is applied, adding the input to the output.
%-------------------------------------------------------------------------
\begin{table*}[ht!]
\footnotesize
\begin{center}
\setlength{\tabcolsep}{1pt}
\begin{tabular}
{p{0.15\linewidth}P{0.16\linewidth}P{0.16\linewidth}P{0.16\linewidth}P{0.16\linewidth}P{0.16\linewidth}}
    \toprule
    Category &S-T\cite{bergmann2020uninformed} &PatchCore\cite{roth2022towards} &GCAD\cite{bergmann2022beyond} &EfficientAD\cite{batzner2023efficientad} &SPACE\\
    \midrule
    \midrule
    Candle &(84.4, 98.9) &(\textbf{98.6}, 98.9) &(79.2, 96.3) &(98.4, 99.1) &(97.2, \textbf{99.5})\\
    Capsules &(85.7, 99.2) &(76.4, 83.6) &(72.4, 91.9) &(\textbf{93.5}, 98.2) &(92.9, \textbf{99.4})\\
    Cashew &(93.7, 96.6) &(97.9, \textbf{99.2}) &(97.5, 98.5) &(97.2, \textbf{99.2}) &(\textbf{98.0}, 99.1)\\
    Chewing gum &(99.1, \textbf{99.1}) &(98.9, 98.1) &(97.3, 98.3) &(99.9, 99.2) &(\textbf{100}, 98.6)\\
    Fryum &(96.8, 96.1) &(94.8, 95.9) &(84.3, \textbf{97.2}) &(96.5, 96.5) &(\textbf{98.9}, 96.5)\\
    Macaroni 1 &(97.7, 99.8) &(95.8, 90.0) &(78.6, 99.3) &(99.4, \textbf{99.9}) &(\textbf{100}, \textbf{99.9})\\
    Macaroni 2 &(88.1, 99.6) &(77.7, 91.0) &(56.3, 97.0) &(\textbf{96.7}, \textbf{99.8}) &(96.2, \textbf{99.8})\\
    PCB 1 &(97.3, 99.4) &(98.9, 99.5) &(95.1, 99.4) &(98.5, 99.3) &(\textbf{99.5}, \textbf{99.8})\\
    PCB 2 &(98.6, 97.1) &(97.1, 97.5) &(98.6, 97.9) &(99.5, \textbf{99.3}) &(\textbf{100}, 98.9)\\
    PCB 3 &(96.4, 98.9) &(96.3, 95.4) &(89.6, 98.7) &(98.9, \textbf{99.4}) &(\textbf{99.7}, 99.3)\\
    PCB 4 &(98.9, 99.1) &(99.4, 98.0) &(97.9, 98.3) &(98.9, 99.1) &(\textbf{99.8}, \textbf{99.2})\\
    Pipe fryum &(98.7, 98.9) &(99.7, \textbf{99.3}) &(97.7, 99.1) &(99.7, \textbf{99.3}) &(\textbf{99.9}, 99.1)\\
    \midrule
    Mean &(94.2 ,98.5) &(93.8, 94.3) &(86.1, 97.5) &(98.1, \textbf{99.1}) &(\textbf{98.5}, \textbf{99.1})\\ 
    \bottomrule
\end{tabular}
\end{center}
\vspace{-0.5cm}
\caption{Detection and localization results on the VisA dataset, using Image-level AUROC(\%) and Pixel-level AUROC(\%). The highest value is represented in bold.}
\label{tab:visa}
\end{table*}
%-------------------------------------------------------------------------
\begin{table*}[t]
\footnotesize
\begin{center}
\setlength{\tabcolsep}{1pt}
\renewcommand{\arraystretch}{1.}
\begin{tabular}
{P{0.1\linewidth}P{0.1\linewidth}P{0.1\linewidth}P{0.1\linewidth}P{0.1\linewidth}P{0.01\linewidth}P{0.1\linewidth}P{0.1\linewidth}P{0.1\linewidth}P{0.1\linewidth}}
    \toprule
    \multirow{3}[2]{*}[-7pt]{Method} &\multicolumn{4}{c}{Components} &&\multicolumn{4}{c}{Performances (AUROC $\%$)}\\
    \cmidrule{2-5}\cmidrule{7-10}
    &\multicolumn{3}{c}{SCL} &\multirow{2}[2]{*}[-3pt]{FM} &&\multirow{2}[2]{*}[-3pt]{Structural} &\multirow{2}[2]{*}[-3pt]{Logical} &\multirow{2}[2]{*}[-3pt]{Total} &\multirow{2}[2]{*}[-3pt]{diff.}\\
    \cmidrule{2-4}
    &\shortstack{Strong\\Branch} &\shortstack{Weak\\Branch} &\shortstack{Selective\\Distillation} & &&&&&\\
    \midrule
    \midrule
    SPACE &\cmark &\cmark &\cmark &\cmark &&95.3 &89.8 &\textbf{92.6} &\textbf{-}\\
    \midrule
    \multirow{4}[2]{*}[1pt]{SCL Ablation} &\xmark &\xmark &\xmark &\cmark &&90.9&88.2&\underline{89.6} &\underline{-3.0}\\
    &\xmark &\cmark &\cmark &\cmark &&91.9 &87.6 &89.7 &-2.8\\
    &\cmark &\xmark &\cmark &\cmark &&93.8 &89.2 &91.5 &-1.1\\
    &\cmark &\cmark &\xmark &\cmark &&93.1 &89.4 &91.3 &-1.3\\
    \midrule
    {FM Ablation} &\cmark &\cmark &\cmark &\xmark &&94.4 &86.8 &\underline{90.6} &\underline{-2.0}\\
    \bottomrule
\end{tabular}
\end{center}
\vspace{-0.5cm}
\caption{\textbf{Key components of the SPACE on MVTec LOCO:}
The results of various components are displayed. The second to fifth rows are related to the \cref{eq:l_obj}. The last row represents the result of removing the FM. The values of the main components are underlined.}
\vspace{-0.25cm}
\label{tab:ablation1}
\end{table*}

The input size is resized at 256$\times$256 dimensions, to leverage SCL, we apply weak augmentation with a random shift of up to 3 pixels. For strong augmentation, we employ both horizontal and vertical image flips, combined with RandAugment~\cite{Cubuk_2020_CVPR_Workshops}, where the parameters $n$ and $m$ are set to 4 and 10, respectively.
The loss balancing parameters $\lambda_1$ and $\lambda_2$ are set as 1 and 0.1, respectively.
And the smoothing factor $\alpha$ for $ema$ is set as 0.999.
We provide more detailed implementation details in the supplementary.

Additionally, SPACE is applicable not only to EfficientAD but also to any CNNs, and the experimental results are provided in the ablation studies.

%-------------------------------------------------------------------------
\subsection{Comparison to State-Of-The-Art Methods}\label{sec:comparison}
In this section, we compared our approach to various methods. \cref{tab:experiment_01} presents the performance comparison results of SPACE and other recent methods on MVTec LOCO.
Our method outperformed all other detection methods, and achieved the second-highest localization sPRO performance, next to GCAD.
Unlike conventional AD, which identifies precise defect objects, MVTec LOCO is designed to predict the existence of objects, including their positions and whether spaces are empty or filled.
For example, if a splicing connector is missing, the ground truth mask for the defect is the entire image. The metric for sPRO counts any area above a certain size as detected, but our method tends to create a finer and more detailed anomaly map that usually cannot exceed this threshold.
As shown in \cref{tab:experiment_02} and \cref{tab:visa}, our approach outperforms recent various methods in both image-level and pixel-level detection on MVTec AD and VisA. Notably, on MVTec AD, our method achieved AUROC of 99.2$\%$ in image-level and 98.8$\%$ in pixel-level detection. In the case of VisA, SPACE reached AUROC of 98.5$\%$ for image-level and 99.1$\%$ for pixel-level detection.
\subsection{Ablation Study}\label{sec:ablation}
\begin{figure}[t]
    \centering
    \resizebox{1.0\columnwidth}{!}{
    \setlength{\tabcolsep}{1pt}
    \begin{tabular}{ccccc}
        \includegraphics[width=\columnwidth]{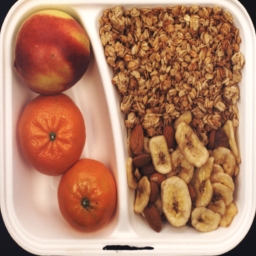} &\includegraphics[width=\columnwidth]{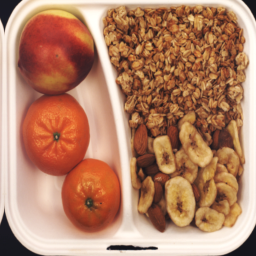} 
        &\includegraphics[width=\columnwidth]{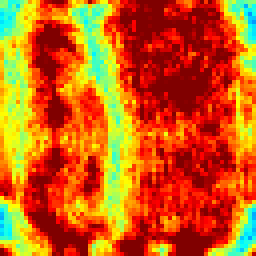} &\includegraphics[width=\columnwidth]{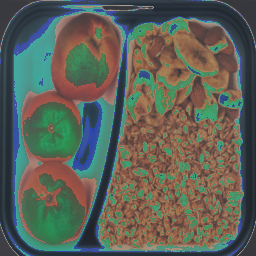} &\includegraphics[width=\columnwidth]{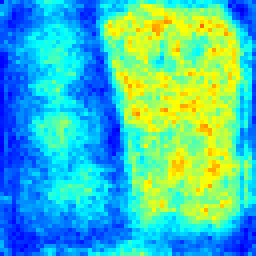}\\
        \includegraphics[width=\columnwidth]{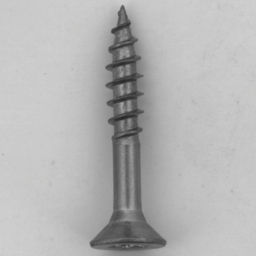} &\includegraphics[width=\columnwidth]{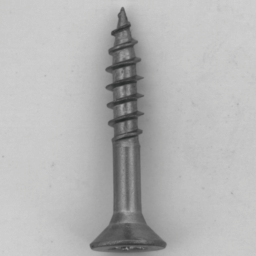} 
        &\includegraphics[width=\columnwidth]{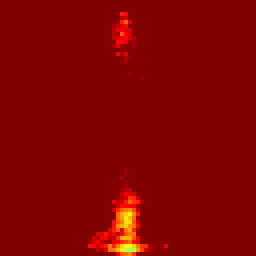} &\includegraphics[width=\columnwidth]{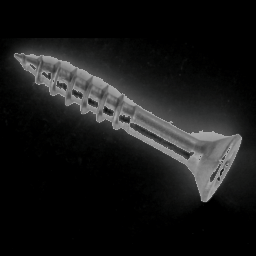} &\includegraphics[width=\columnwidth]{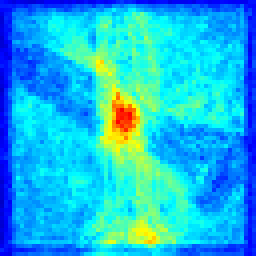}\\             
        \\\fontsize{1.5cm}{1.5cm}\selectfont{(a) Original} &\fontsize{1.5cm}{1.5cm}\selectfont{(b) W. aug.} &\fontsize{1.5cm}{1.5cm}\selectfont{(c) W. region} &\fontsize{1.5cm}{1.5cm}\selectfont{(d) S. aug.} &\fontsize{1.5cm}{1.5cm}\selectfont{(e) S. region}
    \end{tabular}}
    \vspace{-0.05cm}
    \caption{\textbf{The region for updating features in augmented images:}
    In (c) and (e), the color scale signifies that a shift toward red corresponds to a larger number of channels employed for learning in the feature, while a shift toward blue indicates the use of fewer channels.}
    \label{fig:ablation2} 
    \vspace{-0.5cm}
\end{figure}

%-------------------------------------------------------------------------
\noindent\textbf{Effectiveness of SCL:}
SCL is one of the crucial components of our proposed method. This component aims to enhance the detection of structural anomalies by employing both weak and strong augmentations. To confirm the effectiveness of SCL, we conducted ablation studies, and the results are presented in \cref{tab:ablation1}.
The second row shows the results of removing all SCL loss terms, which resulted in a significant performance decrease of 3.0\%. 
The third row shows the results of removing the loss terms $\mathcal{L}_{os}$ and $\mathcal{L}_{ws}$ related to the strong branch in SCL, which led to a performance decrease of 2.8\%. Similarly, the fourth row presents the results of removing the loss terms related to the weak branch. The fifth row presented the results of removing the selected mask $\Upsilon$, which is included in the terms $\mathcal{L}_{ts}$. In this manner, we confirm that SCL has a more pronounced effect on detecting structural abnormalities
Additionally, strong augmentation emerges as the crucial factor within SCL components. The performance is improved by selective distillation between $g_s$ and $g_t$, emphasizing the necessity of feature differences between them.
Especially, \cref{eq:l_ts} ensures that $g_s$ learns only features with differences exceeding a predefined threshold, rather than copying all the features of $g_t$.
Also, \cref{fig:ablation2} illustrates where SCL is focusing its learning for both weak and strong augmentation images. It is evident that SCL effectively identifies and utilizes features in images that are similar to the normal ones. As a result, our proposed loss enables the utilization of various
augmentations for anomaly detection training.

\begin{figure}[t]
    \centering
    \resizebox{0.9\columnwidth}{!}{
    \setlength{\tabcolsep}{1pt}
    \begin{tabular}{cccc}
        \includegraphics[width=\columnwidth]{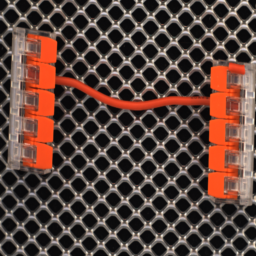}
        &\includegraphics[width=\columnwidth]{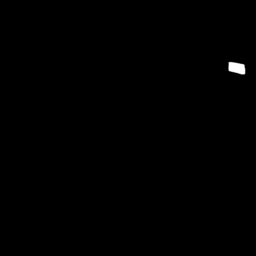} 
        &\includegraphics[width=\columnwidth]{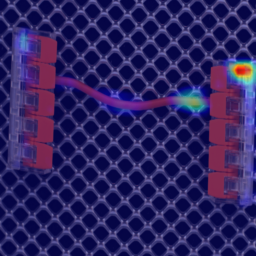}
        &\includegraphics[width=\columnwidth]{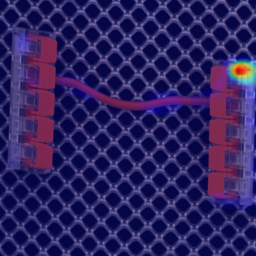}\\        
    \end{tabular}}
    \vspace{-0.25cm}
    \caption{\textbf{Differences in structural anomaly maps when using FM:} From left to right, the sequence represents the original image, the ground truth mask, the anomaly detection map without FM, and the anomaly detection map with FM.}
    \label{fig:ablation_FM1} 
    \vspace{-0.25cm}
\end{figure}

\begin{figure}[t]
    \centering
    \resizebox{0.9\columnwidth}{!}{
    \setlength{\tabcolsep}{1pt}
    \begin{tabular}{cccc}
        \includegraphics[width=\columnwidth]{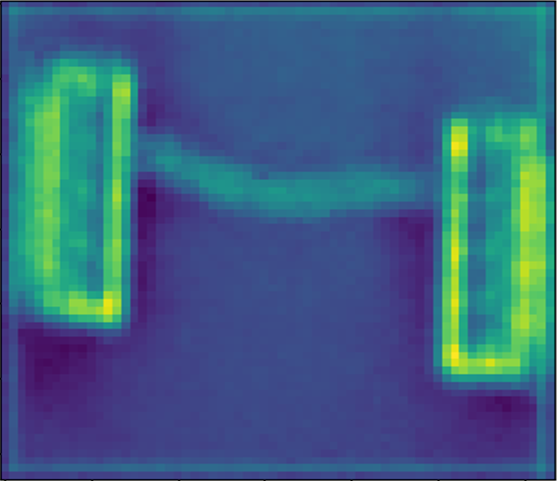}
        &\includegraphics[width=\columnwidth]{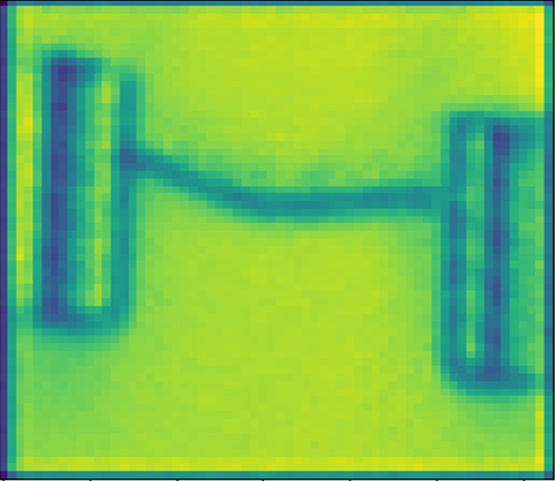} 
        &\includegraphics[width=\columnwidth]{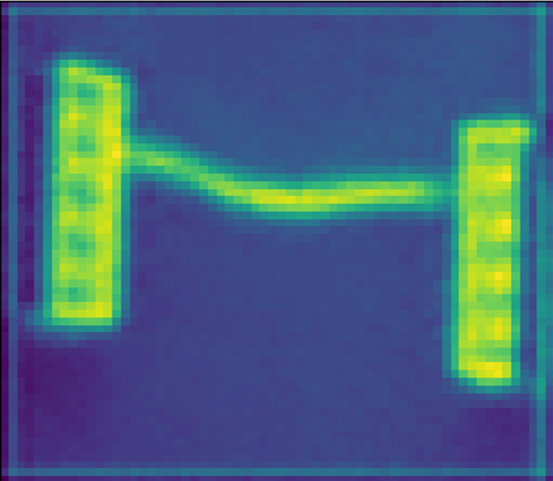}
        &\includegraphics[width=\columnwidth]{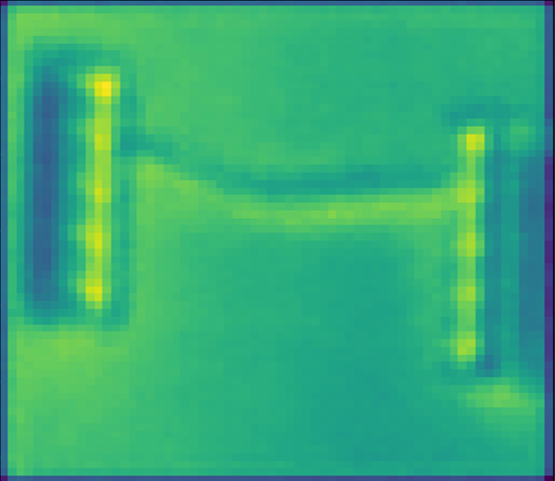}\\
        \includegraphics[width=\columnwidth]{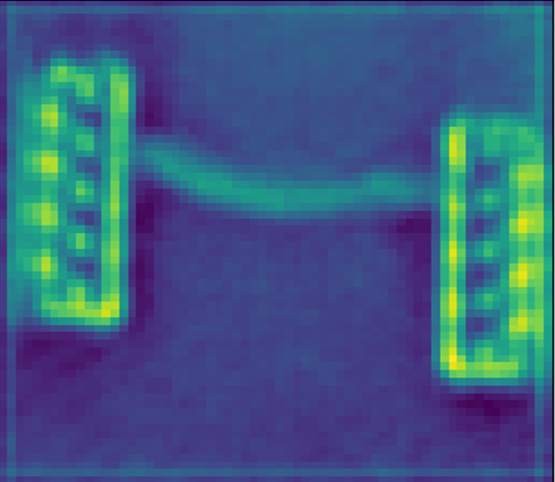}
        &\includegraphics[width=\columnwidth]{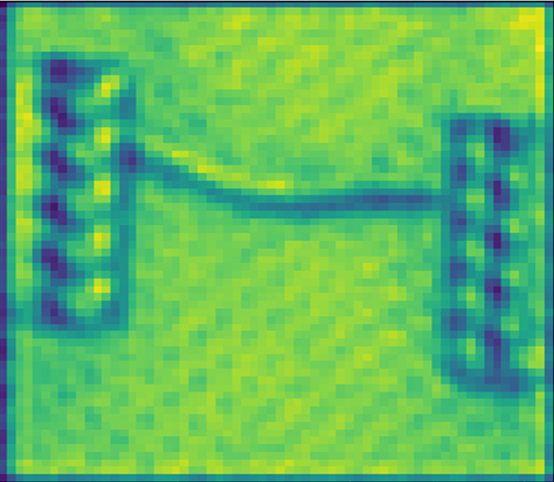} 
        &\includegraphics[width=\columnwidth]{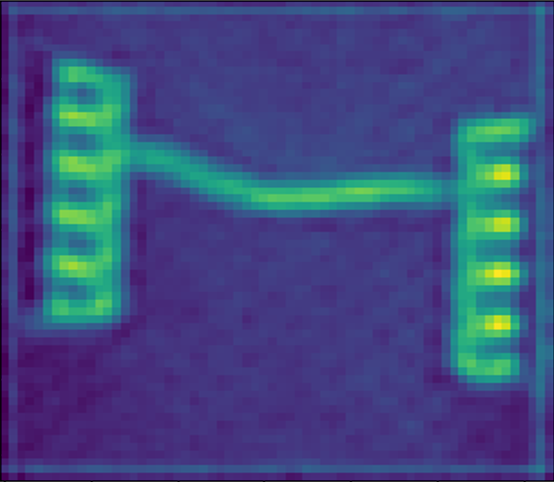}
        &\includegraphics[width=\columnwidth]{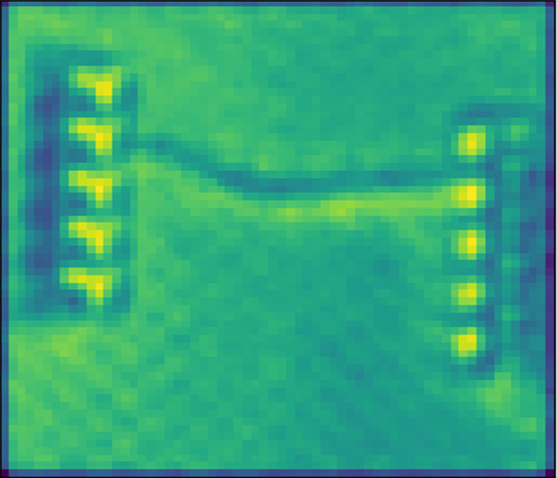}\\        
    \end{tabular}}
    \vspace{-0.25cm}
    \caption{\textbf{Comparison of feature maps without and with FM:} The top row shows the feature map of the student model without FM, while the bottom row shows the feature map of the student model with FM.}
    \label{fig:ablation_FM2} 
    \vspace{-0.25cm}
\end{figure}

%-------------------------------------------------------------------------
\noindent\textbf{Effectiveness of FM:}
FM is another key component of our method. Through this component, SPACE is able to detect logical anomalies. To confirm the effectiveness of FM, we also conducted ablation studies, and the results are shown in \cref{tab:ablation1}. The results in the last row depict the outcome of removing the FM, indicating that $g_s$ allocates half of its final layer to detect logical abnormalities. This adjustment led to a 2\% decrease in performance.
Also, we observed that the removal of FM leads to a 0.9$\%$ decrease in structural anomaly performance. 
It means that FM effectively prevents side effects that degrade the detection of structural anomalies by preventing the blurry features of FE from affecting the student.
To confirm the effect, we compare the anomaly detection maps when FM is used and when it is not. \cref{fig:ablation_FM1} and \cref{fig:ablation_FM2} illustrate the structural anomaly detection maps and features with and without FM. Clearly, when FM is applied, we observe sharper anomaly detection maps. This indicates that FM prevents the influence of blurry information from FE.
%-------------------------------------------------------------------------

%-------------------------------------------------------------------------
%-------------------------------------------------------------------------
% ------------------------------------------------------------------------------------
\begin{table}[h!]
\footnotesize
\begin{center}
\setlength{\tabcolsep}{1pt}
\renewcommand{\arraystretch}{1.0}
\begin{tabular}
    {p{0.22\columnwidth}P{0.2\columnwidth}P{0.17\columnwidth}P{0.17\columnwidth}P{0.17\columnwidth}}
    \toprule
    \multirow{2}[2]{*}[0pt]{Method} &\multirow{2}[2]{*}[0pt]{Pre-trained} &\multicolumn{2}{c}{Network} &\multirow{2}[2]{*}[0pt]{AUROC (\%)}\\
    \cmidrule{3-4}
    &&Teacher &Student &\\
    \midrule
    \midrule
    EfficientAD\cite{batzner2023efficientad} &ImageNet 1K &PDN-M &PDN-M &90.6\\
    \midrule
    \multirow{5}[2]{*}[-4pt]{SPACE} &\multirow{4}[2]{*}[3pt]{ImageNet 1K} &PDN-M &PDN-M &92.6\\
    & &PDN-M &PDN-M$\times$2 &92.8\\
    & &WRN-101 &PDN-M &93.1\\
    & &PDN-S &PDN-S &91.8\\
    \cmidrule{2-5}
    & ImageNet 21K &Resnet50 &PDN-M &93.8\\
    \cmidrule{2-5}
    & CLIP-based &Resnet50 &PDN-M &87.5\\
    \bottomrule
\end{tabular}
\vspace{-0.5cm}
\end{center}
\caption{The performances based on the networks in MVTec LOCO. PDN$\times$2 indicates using twice the number of features.}
\label{tab:network}
\vspace{-0.25cm}
\end{table}

% ------------------------------------------------------------------------------------
\noindent\textbf{Analysis of Networks:}
\cref{tab:network} shows the difference in image-level AUROC performance based on various combinations. 
In PDN, the structure is distilled using the 2nd and 3rd outputs of the residual blocks from WRN-101.
But,
for both WRN-101 and ResNet-50, the outputs of the residual blocks were used directly without distillation in a smaller structure. 
WRN-101 refers to a model pre-trained using ImageNet 1K, while ResNet-50 refers to a model pre-trained using a larger dataset, ImageNet 21K, along with two models utilizing the image encoder of CLIP pre-trained~\cite{radford2021learning}.

Ultimately, rows 2 and 4 show that as the size of the teacher increases, indicating greater diversity in teacher characteristics, the performance of anomaly detection improves.
Furthermore, an interesting observation was that even when a smaller PDN was used as the teacher, increasing the feature size of the student by a factor of two led to performance improvement. This can be attributed to SCL augmenting the data and leveraging more diverse features. An interesting outcome is evident in the final row of the results. Contrary to the generally superior performance of pre-trained with CLIP in various tasks~\cite{zhu2023learning, mokady2021clipcap, deng2023prompt}, in anomaly detection, models with a teacher trained on classification tasks outperformed those with a CLIP-based. In anomaly detection, features learned through definitive labeling in classification might be more effective than those learned through contrastive methods like CLIP. 
%-------------------------------------------------------------------------
\begin{table}[t]
\footnotesize
\begin{center}
\setlength{\tabcolsep}{1pt}
\renewcommand{\arraystretch}{1.0}
\begin{tabular}
{p{0.25\linewidth}P{0.23\linewidth}P{0.23\linewidth}P{0.23\linewidth}}
    \toprule
    \multirow{2}[2]{*}[1pt]{~Category} &\multicolumn{2}{c}{EfficientAD\cite{batzner2023efficientad}} &\multirow{2}[2]{*}[1pt]{Ours}\\
    \cmidrule{2-3}
    &Vanilla & + RandAug & \\
    \midrule
    \midrule
    Breakfast &\textbf{87.5} &75.2 &\textbf{87.5}\\
    Bottle &99.5 &93.9 &\textbf{10}0\\
    Pushpins &96.4 &81.1 &\textbf{98.8}\\
    Screw bag &74.3 &63.2 &\textbf{78.3}\\
    Connector &95.3 &80.4 &\textbf{98.2}\\
    \midrule
    Mean &90.6 &80.4 &\textbf{92.6}\\
    \bottomrule
\end{tabular}
\end{center}
\vspace{-0.25cm}
\caption{The performances are due to strong augmentations.}
\label{tab:strong_aug}
\vspace{-0.25cm}
\end{table}
% ---------------------------------------------------------------
\begin{figure}[t]
    \centering
    \resizebox{.8\columnwidth}{!}{
    \setlength{\tabcolsep}{1pt}
    \begin{tabular}{c}
        \includegraphics[width=\columnwidth, trim=0cm 0cm 0cm 0.5cm]{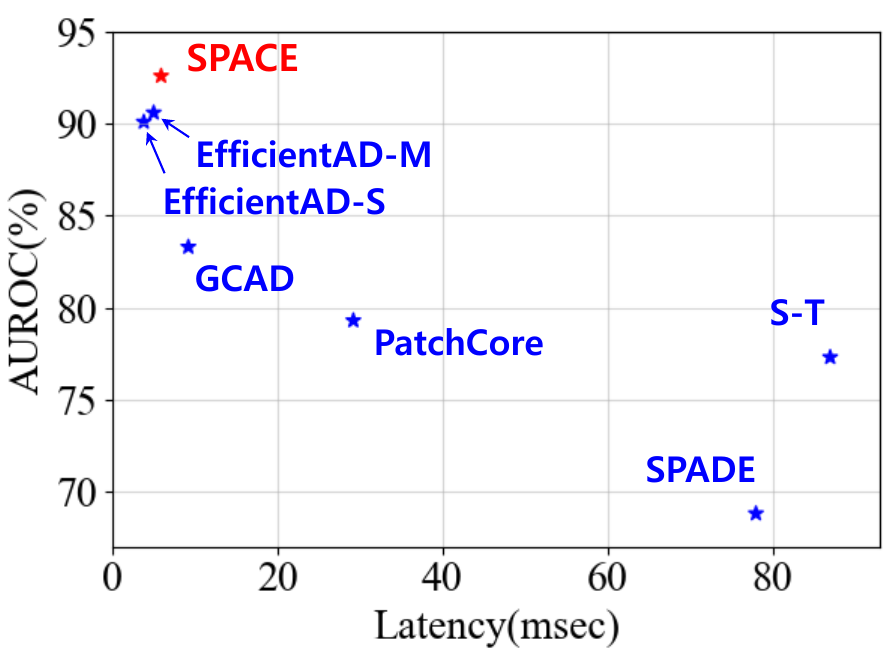}
    \end{tabular}}
    \vspace{-0.25cm}
    \caption{Latency vs. Detection performance on an RTX 3090.}
    \label{fig:ablation_time}
    \vspace{-0.25cm}
\end{figure}

%-------------------------------------------------------------------------
\noindent\textbf{Strong Augmentation:}
This study aims to validate our claim that strong augmentations cannot be effectively utilized in AD. To this end, we conducted experiments applying RandAugment~\cite{Cubuk_2020_CVPR_Workshops} to the state-of-the-arts, EfficientAD. The results of these experiments are shown in~\cref{tab:strong_aug}. As we argued, simply applying strong augmentation negatively impacts the performance of anomaly detection models. In contrast, our proposed method, SCL, demonstrates the ability to selectively learn useful features even when strong augmentation is applied. These findings support our claim and suggest that SCL can effectively leverage strong augmentation in the context of anomaly detection.

%-------------------------------------------------------------------------
\noindent\textbf{Analysis of Latency and Detection:}
In real-world industrial environments, inference time is a key factor. We compare latency and anomaly detection performance with other methods in \cref{fig:ablation_time}. The comparison dataset used is MVTec LOCO, and all tests were conducted
using the RTX 3090. As a result, our method demonstrates the highest performance in terms of AUROC, while maintaining a similar latency to EfficientAD, which currently shows the fastest inference time.
Also, SPACE demonstrates better performances, even with a lightweight model designed for low latency.
In \cref{tab:mvtecad}, it seems lower than the state-of-the-arts, ReconPatch~\cite{Hyun_2024_WACV}, which is based on a heavy-weight model, WRN-50. However, as shown in \cref{fig:ablation_time}, we aim to improve both latency and detection performance, so we adopted the lightweight model, PDN. Focusing on detection performance, we can also adopt ResNet-50, and it shows better results than the state-of-the-art methods.
%-------------------------------------------------------------------------
\begin{table}[t]
\begin{center}
\footnotesize
\setlength{\tabcolsep}{1.pt}
\renewcommand{\arraystretch}{1.0}
\begin{tabular}
    {p{0.22\columnwidth}P{0.2\columnwidth}P{0.18\columnwidth}P{0.17\columnwidth}P{0.17\columnwidth}}
    \toprule
    \multirow{2}[2]{*}[0pt]{Method} &\multirow{2}[2]{*}[0pt]{Network} &\multicolumn{3}{c}{AUROC (\%)}\\
    \cmidrule{3-5}
    &&Texture Avg. &Object Avg. &Total Avg\\
    \midrule
    \midrule
    ReconPatch~\cite{Hyun_2024_WACV} &WRN-50 &99.8 &99.4 &99.6\\
    \midrule
    \multirow{2}[2]{*}[4pt]{SPACE} &PDN &99.8 &99.0 &99.2\\
    &ResNet-50 &99.9 &99.8 &99.7\\
    \bottomrule
\end{tabular}
\end{center}
\vspace{-0.25cm}
\caption{Performances according to the Teachers in MVTecAD.}
\label{tab:mvtecad}
\vspace{-0.5cm}
\end{table}

%-------------------------------------------------------------------------
\subsection{Limitations}\label{sec:limitations}
Our model selects updating features using \cref{eq:upsilon}, which performs a similar role to the confidence threshold in semi-supervised learning.
This equation essentially calculates the difference between
$g_t$ and $g_s$
, takes a moving average, and provides a rough estimate of the mean at each location. Therefore, if we can establish a more precise criterion to select features from augmented images, we could achieve better performance. For instance, in semi-supervised learning, ConMatch~\cite{kim2022conmatch} introduces a confidence network to use more accurate pseudo-labels. 
Adopting a learnable approach for selecting more precise features has the potential to enhance performance.
%-------------------------------------------------------------------------
\section{Conclusion}\label{sec:conclusion}
We introduced a novel unsupervised anomaly detection method with spatial-aware consistency regularization to overcome distribution shifts between natural and industrial datasets.
The approach improved detection performance by learning target-oriented feature representations through consistency regularization, utilizing augmented normal patterns without the need for specific labels. 
Moreover, the proposed feature converter module protects the structural detection of the student model by aligning with the blurred features from the feature-encoder, minimizing false positives, and consequently sharpening the differentiation between normal and abnormal features.
Applying this method, the anomaly detection AUROC achieved 92.6$\%$, 99.2$\%$, and 98.5$\%$ on the widely validated MVTec LOCO, MVTec AD, and VisA datasets, respectively, demonstrating its effectiveness in learning industrial normal features.
Although SPACE exploits a straightforward yet stable criterion updated by $ema$, we anticipate further performance improvements by employing sophisticated criteria that consider both logical and structural branches and select appropriate features. Our future work will focus on exploring advanced criteria to refine and optimize the proposed method.

%-------------------------------------------------------------------------
\section{Acknowledgements}
This work was funded by Samsung Electro-Mechanics and was partially supported by Carl-Zeiss Stiftung under the Sustainable Embedded AI project (P2021-02-009).

%-------------------------------------------------------------------------
%%%%%%%%% REFERENCES
{\small
\bibliographystyle{ieee_fullname}
\bibliography{egbib}
}

%-------------------------------------------------------------------------
\clearpage
\appendix

\renewcommand{\thetable}{\Alph{table}}
\renewcommand{\thefigure}{\Alph{figure}}
\renewcommand{\thealgorithm}{\Alph{algorithm}}
\renewcommand{\thesection}{\Alph{section}}

% ---------------------------------------------------------------
\begin{algorithm}[!ht]
\caption{Spatial-aware Consistency Loss (SCL)}
\label{alg:maxValue}
\small
\begin{spacing}{1.25}
\begin{algorithmic}[1]
\STATE \textbf{{Notation:} teacher network $g_t$, student network $g_s$\\}
\STATE \textbf{{Input:} original image $x_o$, weak augmented image $x_w$, strong augmented image $x_s$\\}
\STATE
\IF{iteration $<$ 5000}
\STATE {$\lambda_1$ = 0.0}\\ 
\ELSE
\STATE {$\lambda_1$ = 1.0}\\
\ENDIF
\STATE
\STATE \textbf{\# Calculate relative distances for masks}
\STATE {$F_{ts}^{o}$ = square($g_t$($x_o$) - $g_s$($x_o$))}\\
\STATE {$F_{ts}^{w}$ = square($g_t$($x_o$) - $g_s$($x_w$))}\\
\STATE {$F_{ts}^{s}$ = square($g_t$($x_o$) - $g_s$($x_s$))}\\
\STATE \textbf{\# Calculate consistency distances in student}
\STATE {$D^{ow}$ = square(stopgrad($g_s$($x_o$)) - $g_s$($x_w$))}\\
\STATE {$D^{os}$ = square(stopgrad($g_s$($x_o$)) - $g_s$($x_s$))}\\
\STATE {$D^{ws}$ = square($g_s$($x_w$) - $g_s$($x_s$))}\\
\STATE \textbf{\# Create update masks}
\STATE {$M^o$ = $F_{ts}^{o}$ $>$ $\Upsilon$}\\
\STATE {$M^w$ = $F_{ts}^{w}$ $<$ $\Upsilon$}\\
\STATE {$M^s$ = $F_{ts}^{s}$ $<$ $\Upsilon$}\\
\STATE \textbf{\# Compute local losses}
\STATE {$\mathcal{L}_{ts}$ = sum($F_{ts}^{o}$ $\odot$ $M^o$) / sum($M^o$)}\\
\STATE {$\mathcal{L}_{unw}$ = sum($D^{os}$ $\odot$ $M^w$) / sum($M^w$)}\\
\STATE {$\mathcal{L}_{uns}$ = sum($D^{os}$ $\odot$ $M^s$) / sum($M^s$)}\\
\STATE {$\mathcal{L}_{unws}$ = sum($D^{ws}$ $\odot$ $M^w$ $\odot$ $M^s$ ) / sum($M^w$ * $M^s$)}\\
\STATE \textbf{\# Update criterion}
\IF{iteration == 0}
\STATE {$\Upsilon$ = $F_{ts}^{o}$}\\ 
\ELSE
\STATE {$\Upsilon$ = $\alpha$~$\times$~$\Upsilon$ + (1 - $\alpha$)~$\times$~$F_{ts}^{o}$}\\
\ENDIF
\STATE
\STATE $\mathcal{L}_{local}$ = $\mathcal{L}_{ts}$ + $\lambda_1$$\times$ ($\mathcal{L}_{unw}$ + $\mathcal{L}_{uns}$ + $\mathcal{L}_{unws}$)\\
\STATE
\RETURN{$\mathcal{L}_{local}$} 
\end{algorithmic}
\end{spacing}
\end{algorithm}
% ---------------------------------------------------------------
\section{Input and Training Details}
Our model has four input branches. Among these, the three inputs are dedicated to the training of the student model, and the other one is for training the feature-encoder and feature-converter module.
For the input size, the original images are resized to 256$\times$256 dimensions. It performs normalization using the mean and standard deviation values from the ImageNet dataset, where the mean values are 0.485, 0.456, and 0.406, and the standard deviations are 0.229, 0.224, and 0.225. Additionally, weak augmentation involves shifting the images randomly by up to 3 pixels, while strong augmentation combines RandAugment~\cite{Cubuk_2020_CVPR_Workshops} with horizontal and vertical flips. The parameters for RandAugment are set to 4 and 10, respectively. For the feature-encoder and Feature-converter Module (FM), the original images are resized to 256$\times$256 size. Furthermore, augmentation is performed, including brightness, contrast, and saturation, each with a parameter value of 0.2.
The training procedure is conducted using the Adam optimizer with a batch size of 1, involving 120,000 iterations on MVTec LOCO and 70,000 iterations on MVTec 2D.

The weight decay is set to 1e-5 for the student model, while 1e-6 is used for both the feature-encoder and FM. Additionally, to align the training speed of the feature-encoder with that of the student model, we update the student weights using an exponential moving average with a parameter of 0.999. The specific Spatial-aware Consistency Loss (SCL) algorithm for the student model is described in~\cref{alg:maxValue}.
% ---------------------------------------------------------------
\section{Analysis of Parameters}
\subsection{Analysis of $\lambda_1$ and $\lambda_2$}\label{sec:parameter1}
We study the impact of the loss balancing parameters $\lambda_1$ and $\lambda_2$. $\lambda_1$ is a parameter associated with SCL, which regulates when the consistency loss is applied. To address the potential low accuracy of the student model before learning, $\lambda_1$ is set at 0 for the initial 5,000 iterations and is later changed to 1.0 after some level of learning has taken place. \cref{tab:lambda1} and \cref{tab:lambda2} presents the changes in image-level AUROC on the MVTec LOCO based on the number of iterations with $\lambda_1$ maintained at 0. It is observed that applying $\lambda_1$ after the first 5,000 iterations yields better performance than applying it from the outset, with a subsequent decrease in performance observed beyond that point. $\lambda_2$ are parameters that control the learning speed of the feature-encoder and FM. At 0.1, it shows the best performance, with performance gradually decreasing as it increases.
\subsection{Analysis of $d_{hard}$}\label{sec:parameter2}
We study the impact of the parameters $d_{hard}$. It is a parameter that removes unnecessary feature positions, selectively training only the loss values that correspond to the top percentile. For example, when using a value like 0.99, $d_{hard}$ is set as the top 99\% of feature difference values as a threshold, updating only those values that exceed this threshold. $d_{hard}$ were set to 0.99, and the experimental results for this parameter are presented in~\cref{tab:d_hard}. The experimental results showed the best performance when $d_{hard}$ was 0.99. This indicates that distilling only the influential values, rather than distilling all areas of the student's features into the feature-encoder, is more helpful for anomaly detection.
%-------------------------------------------------------------------------
\begin{table}[t]
\small
\begin{center}
\setlength{\tabcolsep}{1pt}
\renewcommand{\arraystretch}{1.}
\begin{tabular}
    {p{0.38\columnwidth}P{0.17\columnwidth}P{0.17\columnwidth}p{0.01\columnwidth}P{0.2\columnwidth}}
    \toprule
    \multirow{2}[2]{*}[0pt]{Dataset} &\multicolumn{2}{c}{Iterations} &&\multirow{2}[2]{*}[0pt]{AUROC (\%)}\\
    \cmidrule{2-3}
    &$\lambda_1$=0 &$\lambda_1$=1 &\\
    \midrule
    \midrule
    \multirow{4}[2]{*}[0pt]{MVTec LOCO} &\centering{-} &$\geq$ 0 &&92.2\\
    &$<$ 5,000 &$\geq$ 5,000 &&\textbf{92.6}\\
    &$<$ 10,000 &$\geq$ 10,000 &&91.9\\
    &$<$ 20,000 &$\geq$ 20,000 &&91.8\\
    \bottomrule
\end{tabular}
\end{center}
\vspace{-0.3cm}
\caption{The performance differences depending on $\lambda_1$.}
\label{tab:lambda1}
\vspace{-0.2cm}
\end{table}
%-------------------------------------------------------------------------
\begin{table}[t]
\small
\begin{center}
\setlength{\tabcolsep}{1pt}
\renewcommand{\arraystretch}{1.}
\begin{tabular}
    {p{0.45\columnwidth}P{0.25\columnwidth}P{0.25\columnwidth}}
    \toprule
    Dataset &$\lambda_2$ &{AUROC (\%)}\\
    \midrule
    \midrule
    \multirow{4}[2]{*}[-5pt]{MVTec LOCO} &{0.01} &90.3\\
    &{0.1} &\textbf{92.6}\\
    &{0.2} &92.2\\
    &{0.4} &92.5\\
    &{0.8} &89.8\\
    &{1.0} &89.6\\
    \bottomrule
\end{tabular}
\end{center}
\vspace{-0.3cm}
\caption{The performance differences depending on $\lambda_2$.}
\label{tab:lambda2}
\vspace{-0.5cm}
\end{table}
% ------------------------------------------------------------------------------------
\begin{table}[ht!]
\small
\begin{center}
\setlength{\tabcolsep}{1pt}
\renewcommand{\arraystretch}{1.0}
\begin{tabular}
{p{0.33\linewidth}P{0.3\linewidth}P{0.3\linewidth}}
    \toprule
    Dataset &$d_{hard}$ &AUROC (\%)\\
    \midrule
    \midrule
    \multirow{5}[2]{*}[3pt]{MVTec LOCO} &0.0 &91.2\\
    &0.5 &91.6\\
    &0.9 &92.0\\
    &0.99 &\textbf{92.6}\\
    &0.999 &92.0\\
    \bottomrule
\end{tabular}
\end{center}
\vspace{-0.3cm}
\caption{The performance differences based on the $d_{hard}$.}
\label{tab:d_hard}
\vspace{-0.5cm}
\end{table}
% ------------------------------------------------------------------------------------
\begin{table}[ht!]
\small
\begin{center}
\setlength{\tabcolsep}{1pt}
\renewcommand{\arraystretch}{1.0}
\begin{tabular}
{p{0.33\linewidth}P{0.3\linewidth}P{0.3\linewidth}}
    \toprule
    Dataset &$ema$ ratio, $\alpha$ &AUROC (\%)\\
    \midrule
    \midrule
    \multirow{5}[2]{*}[3pt]{MVTec LOCO} &0.0 &91.8\\
    &0.5 &91.7\\
    &0.9 &91.3\\
    &0.99 &92.4\\
    &0.999 &\textbf{92.6}\\
    &0.9999 &91.2\\
    \bottomrule
\end{tabular}
\end{center}
\vspace{-0.5cm}
\caption{The performance differences based on the $ema$.}
\label{tab:ema}
\end{table}
% ------------------------------------------------------------------------------------
\begin{figure}[t]
    \centering
    \resizebox{1.\columnwidth}{!}{
    \rotatebox{90}{
    \begin{tabular}{cc}
        \fontsize{1.cm}{1.cm}\selectfont{(b) Capsule}
        &\fontsize{1.cm}{1.cm}\selectfont{(a) Pushpins}\\
        \rotatebox{270}{\includegraphics[width=1.5\columnwidth]{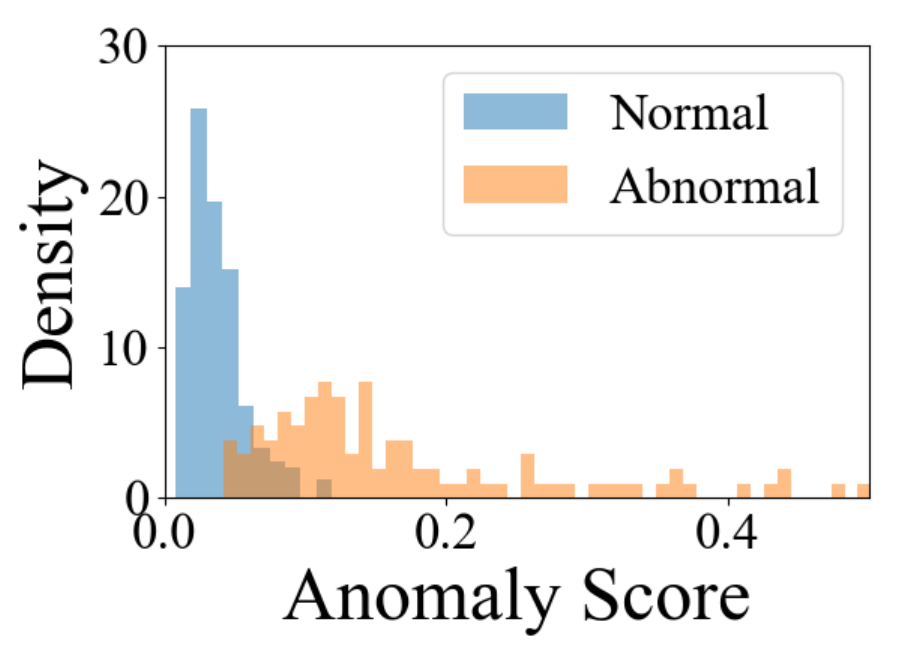}}
        &\rotatebox{270}{\includegraphics[width=1.5\columnwidth]{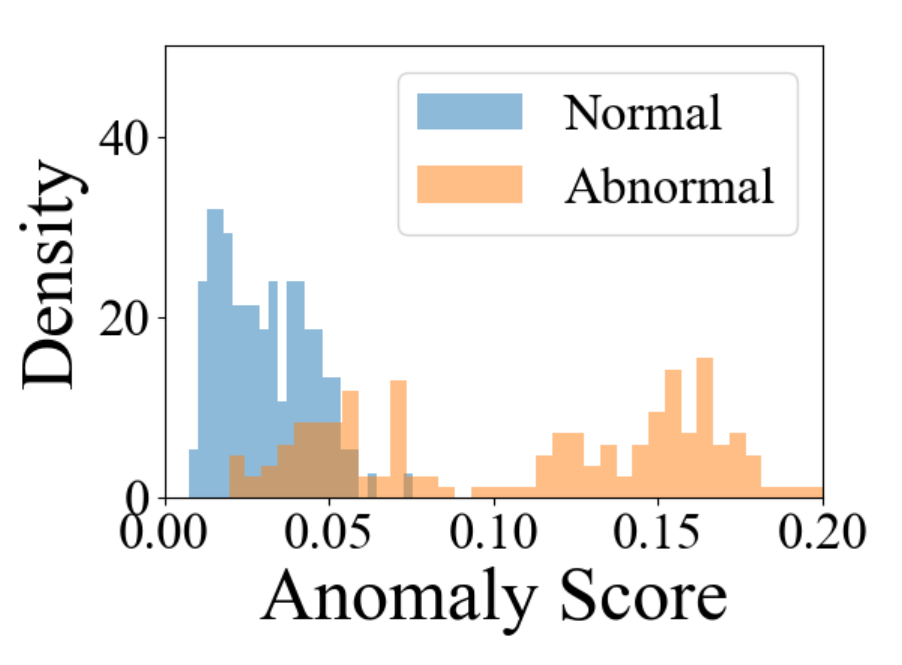}}\\
        \rotatebox{270}{\includegraphics[width=1.5\columnwidth]{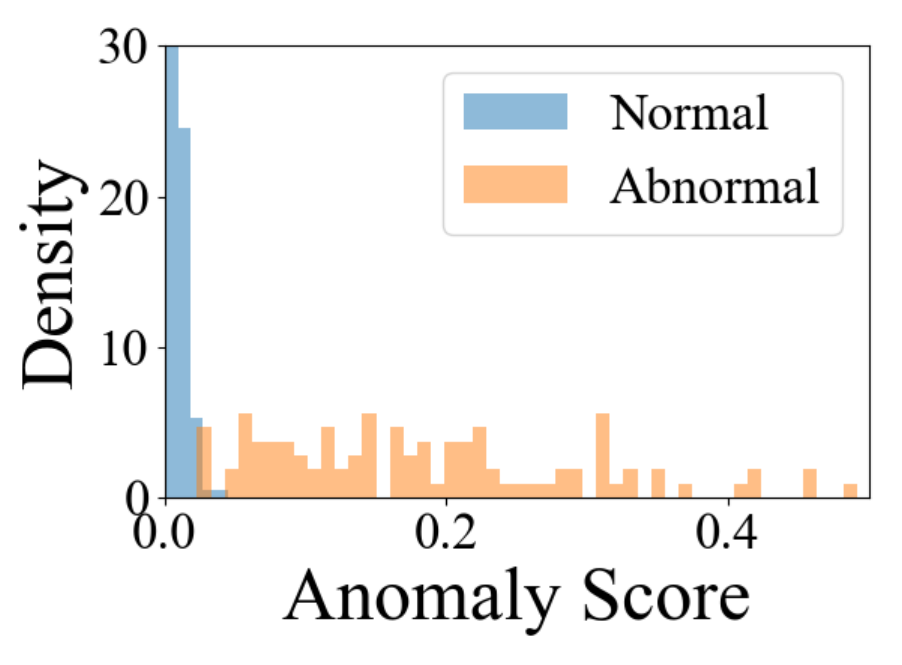}}
        &\rotatebox{270}{\includegraphics[width=1.5\columnwidth]{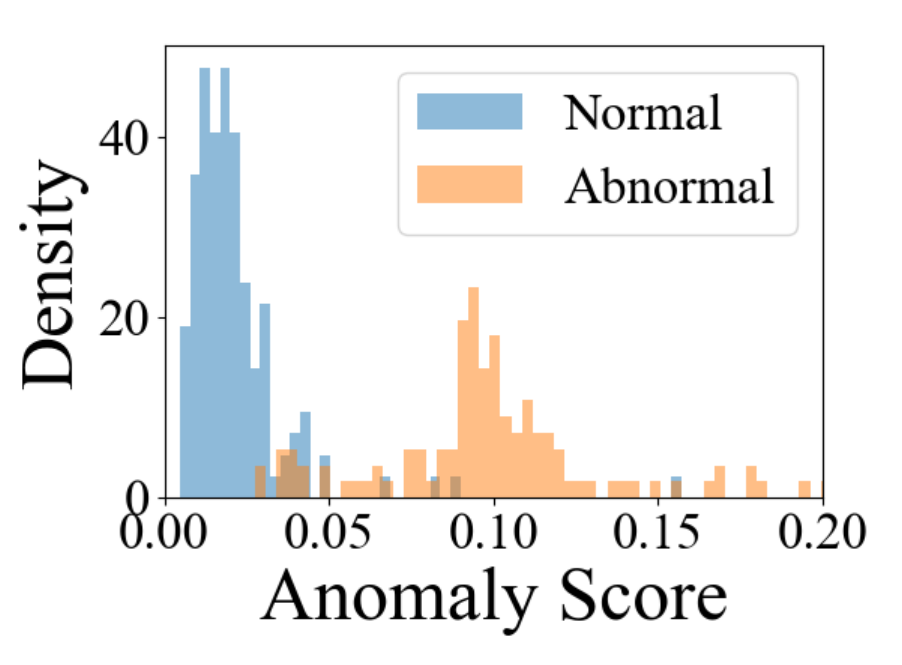}}\\
    \end{tabular}}}
    \caption{\textbf{The histograms of anomaly scores for Pushpins and Capsule:} Left represents the score histogram with only distillation loss, while right shows the histogram with the learning of SCL.} 
    \label{fig:ablation1}
\end{figure}

% ------------------------------------------------------------------------------------
\section{Effectiveness of SCL}
To qualitatively assess the impact of SCL, we conducted additional analysis and \cref{fig:ablation1} presents histograms comparing anomaly scores when SCL is not applied versus when it is applied. As shown, when we utilize the proposed loss, anomaly scores are more distinctly separated, indicating improved discrimination.
% ------------------------------------------------------------------------------------
\section{Effectiveness of EMA}
The effectiveness of utilizing the information necessary for training depends on how the criteria are established. 
We compared performance based on the degree of $ema$, with the results presented in \cref{tab:ema}.
The results indicate that performance varies based on the degree of $ema$.
This indicates that the lightweight ratio is unaffected by SPACE, while a heavyweight ratio exceeding 0.999 is detrimental. However, we confirmed that this value yields the best performance on the evaluated datasets.
% ------------------------------------------------------------------------------------
\section{Qualitative Results}\label{sec:qualitative}
\begin{figure*}[t]
\begin{center}
\begin{subfigure}{.08\textwidth}
\centering
  \includegraphics[width=1.\linewidth]{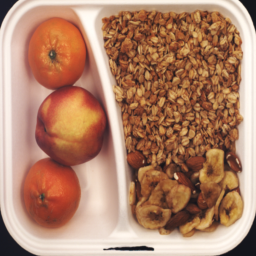}
\end{subfigure}
\begin{subfigure}{.08\textwidth}
\centering
  \includegraphics[width=1.\linewidth]{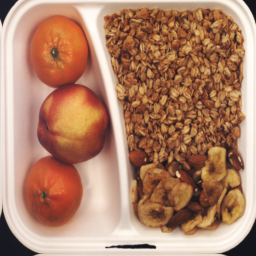}
\end{subfigure}
\begin{subfigure}{.08\textwidth}
\centering
  \includegraphics[width=1.\linewidth]{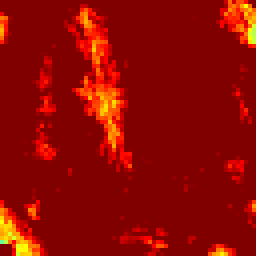}
\end{subfigure}
\begin{subfigure}{.08\textwidth}
\centering
  \includegraphics[width=1.\linewidth]{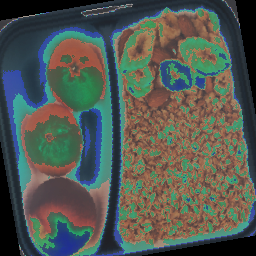}
\end{subfigure}
\begin{subfigure}{.08\textwidth}
\centering
  \includegraphics[width=1.\linewidth]{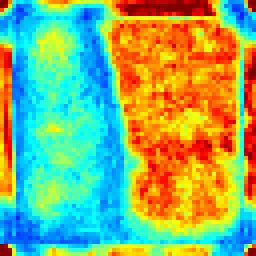}
\end{subfigure}
\begin{subfigure}{.08\textwidth}
\centering
  \includegraphics[width=1.\linewidth]{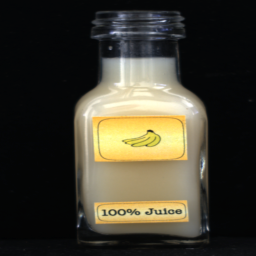}
\end{subfigure}
\begin{subfigure}{.08\textwidth}
\centering
  \includegraphics[width=1.\linewidth]{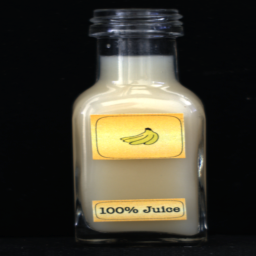}
\end{subfigure}
\begin{subfigure}{.08\textwidth}
\centering
  \includegraphics[width=1.\linewidth]{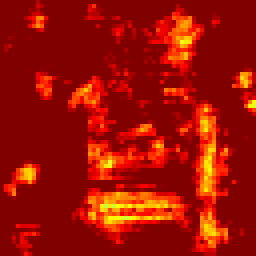}
\end{subfigure}
\begin{subfigure}{.08\textwidth}
\centering
  \includegraphics[width=1.\linewidth]{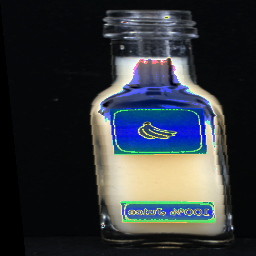}
\end{subfigure}
\begin{subfigure}{.08\textwidth}
\centering
  \includegraphics[width=1.\linewidth]{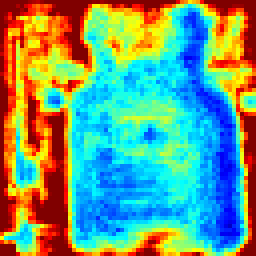}
\end{subfigure}
\vskip\baselineskip
\vspace{-0.5cm}

\begin{subfigure}{.08\textwidth}
\centering
  \includegraphics[width=1.\linewidth]{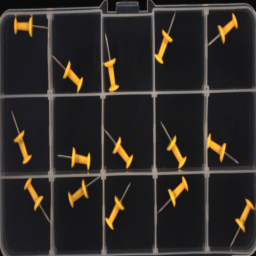}
\end{subfigure}
\begin{subfigure}{.08\textwidth}
\centering
  \includegraphics[width=1.\linewidth]{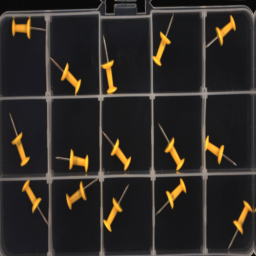}
\end{subfigure}
\begin{subfigure}{.08\textwidth}
\centering
  \includegraphics[width=1.\linewidth]{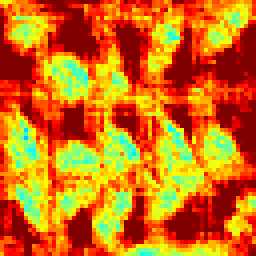}
\end{subfigure}
\begin{subfigure}{.08\textwidth}
\centering
  \includegraphics[width=1.\linewidth]{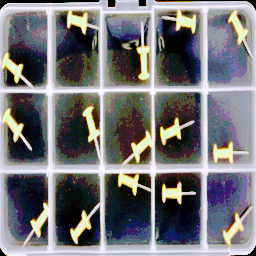}
\end{subfigure}
\begin{subfigure}{.08\textwidth}
\centering
  \includegraphics[width=1.\linewidth]{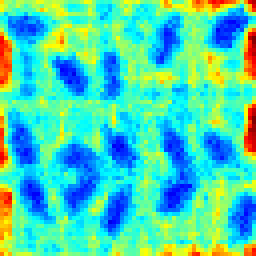}
\end{subfigure}
\begin{subfigure}{.08\textwidth}
\centering
  \includegraphics[width=1.\linewidth]{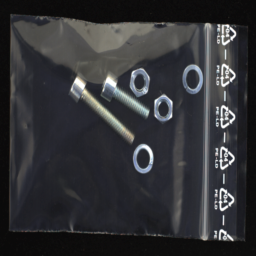}
\end{subfigure}
\begin{subfigure}{.08\textwidth}
\centering
  \includegraphics[width=1.\linewidth]{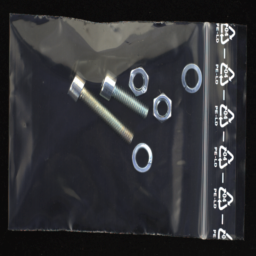}
\end{subfigure}
\begin{subfigure}{.08\textwidth}
\centering
  \includegraphics[width=1.\linewidth]{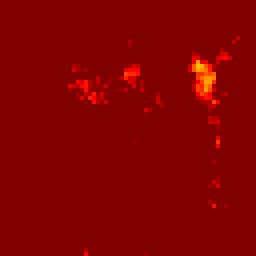}
\end{subfigure}
\begin{subfigure}{.08\textwidth}
\centering
  \includegraphics[width=1.\linewidth]{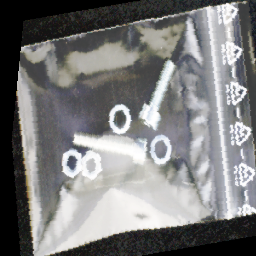}
\end{subfigure}
\begin{subfigure}{.08\textwidth}
\centering
  \includegraphics[width=1.\linewidth]{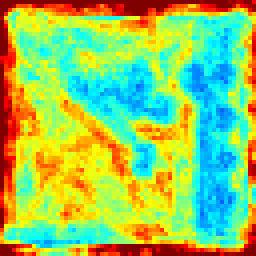}
\end{subfigure}
\vskip\baselineskip
\vspace{-0.5cm}

\begin{subfigure}{.08\textwidth}
\centering
  \includegraphics[width=1.\linewidth]{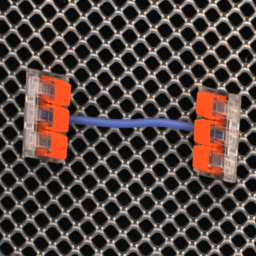}
\end{subfigure}
\begin{subfigure}{.08\textwidth}
\centering
  \includegraphics[width=1.\linewidth]{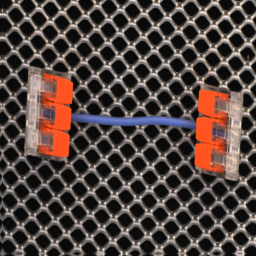}
\end{subfigure}
\begin{subfigure}{.08\textwidth}
\centering
  \includegraphics[width=1.\linewidth]{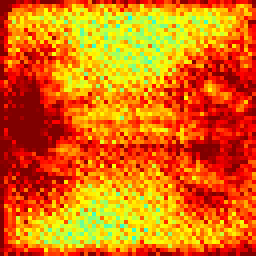}
\end{subfigure}
\begin{subfigure}{.08\textwidth}
\centering
  \includegraphics[width=1.\linewidth]{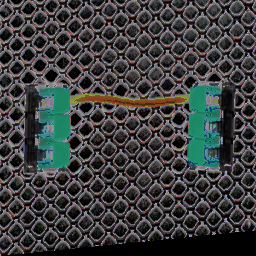}
\end{subfigure}
\begin{subfigure}{.08\textwidth}
\centering
  \includegraphics[width=1.\linewidth]{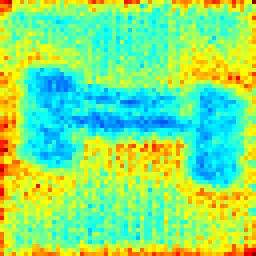}
\end{subfigure}
\begin{subfigure}{.08\textwidth}
\centering
  \includegraphics[width=1.\linewidth]{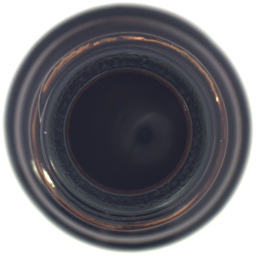}
\end{subfigure}
\begin{subfigure}{.08\textwidth}
\centering
  \includegraphics[width=1.\linewidth]{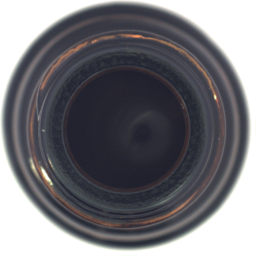}
\end{subfigure}
\begin{subfigure}{.08\textwidth}
\centering
  \includegraphics[width=1.\linewidth]{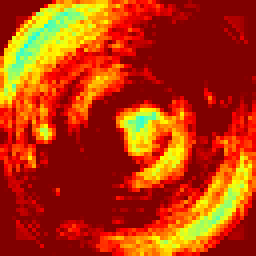}
\end{subfigure}
\begin{subfigure}{.08\textwidth}
\centering
  \includegraphics[width=1.\linewidth]{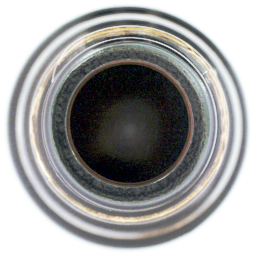}
\end{subfigure}
\begin{subfigure}{.08\textwidth}
\centering
  \includegraphics[width=1.\linewidth]{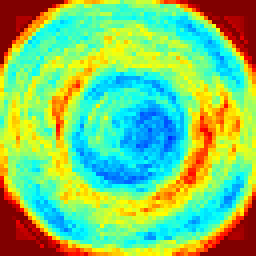}
\end{subfigure}
\vskip\baselineskip
\vspace{-0.5cm}

\begin{subfigure}{.08\textwidth}
\centering
  \includegraphics[width=1.\linewidth]{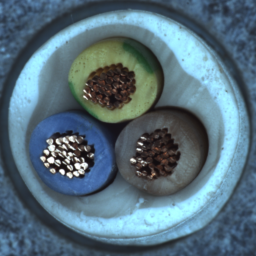}
\end{subfigure}
\begin{subfigure}{.08\textwidth}
\centering
  \includegraphics[width=1.\linewidth]{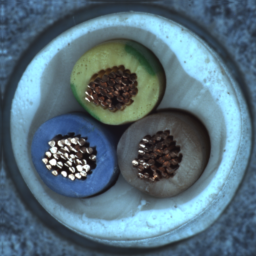}
\end{subfigure}
\begin{subfigure}{.08\textwidth}
\centering
  \includegraphics[width=1.\linewidth]{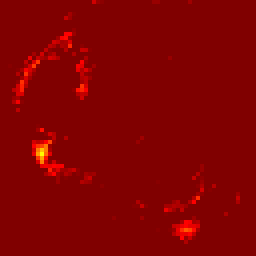}
\end{subfigure}
\begin{subfigure}{.08\textwidth}
\centering
  \includegraphics[width=1.\linewidth]{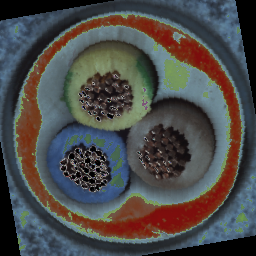}
\end{subfigure}
\begin{subfigure}{.08\textwidth}
\centering
  \includegraphics[width=1.\linewidth]{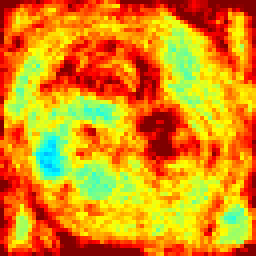}
\end{subfigure}
\begin{subfigure}{.08\textwidth}
\centering
  \includegraphics[width=1.\linewidth]{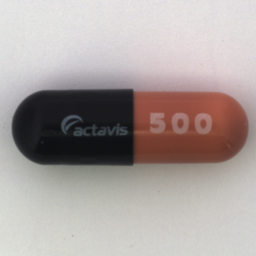}
\end{subfigure}
\begin{subfigure}{.08\textwidth}
\centering
  \includegraphics[width=1.\linewidth]{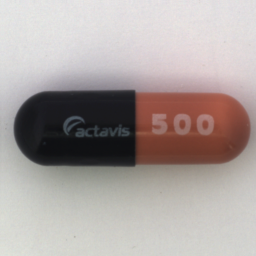}
\end{subfigure}
\begin{subfigure}{.08\textwidth}
\centering
  \includegraphics[width=1.\linewidth]{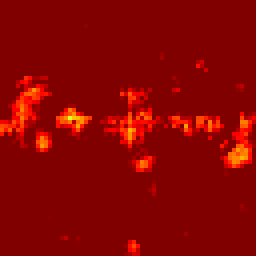}
\end{subfigure}
\begin{subfigure}{.08\textwidth}
\centering
  \includegraphics[width=1.\linewidth]{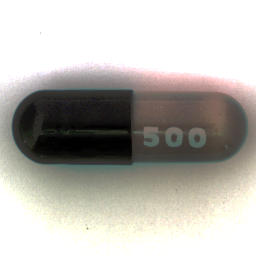}
\end{subfigure}
\begin{subfigure}{.08\textwidth}
\centering
  \includegraphics[width=1.\linewidth]{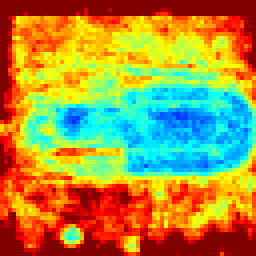}
\end{subfigure}
\vskip\baselineskip
\vspace{-0.5cm}

\begin{subfigure}{.08\textwidth}
\centering
  \includegraphics[width=1.\linewidth]{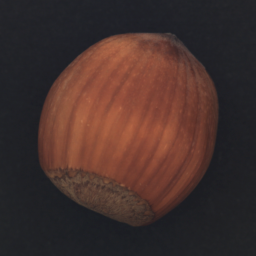}
\end{subfigure}
\begin{subfigure}{.08\textwidth}
\centering
  \includegraphics[width=1.\linewidth]{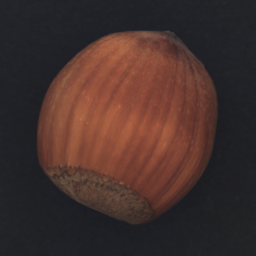}
\end{subfigure}
\begin{subfigure}{.08\textwidth}
\centering
  \includegraphics[width=1.\linewidth]{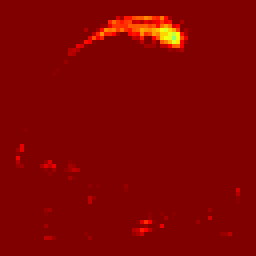}
\end{subfigure}
\begin{subfigure}{.08\textwidth}
\centering
  \includegraphics[width=1.\linewidth]{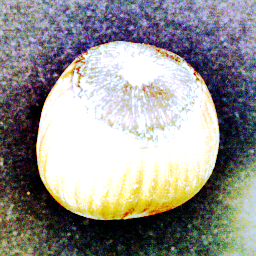}
\end{subfigure}
\begin{subfigure}{.08\textwidth}
\centering
  \includegraphics[width=1.\linewidth]{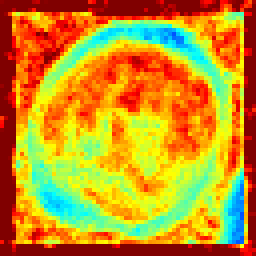}
\end{subfigure}
\begin{subfigure}{.08\textwidth}
\centering
  \includegraphics[width=1.\linewidth]{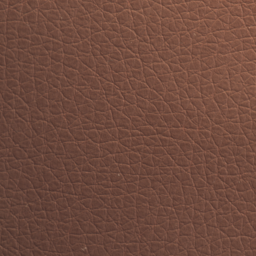}
\end{subfigure}
\begin{subfigure}{.08\textwidth}
\centering
  \includegraphics[width=1.\linewidth]{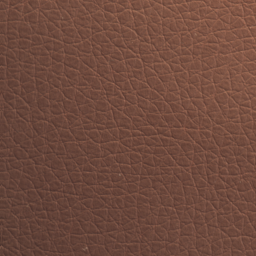}
\end{subfigure}
\begin{subfigure}{.08\textwidth}
\centering
  \includegraphics[width=1.\linewidth]{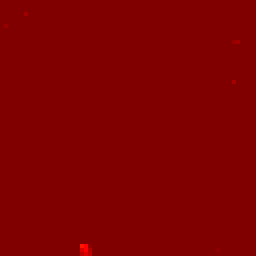}
\end{subfigure}
\begin{subfigure}{.08\textwidth}
\centering
  \includegraphics[width=1.\linewidth]{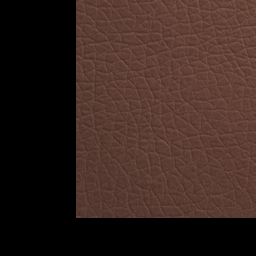}
\end{subfigure}
\begin{subfigure}{.08\textwidth}
\centering
  \includegraphics[width=1.\linewidth]{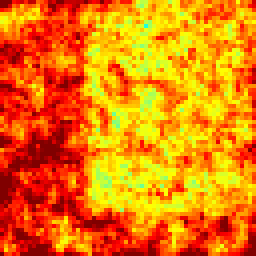}
\end{subfigure}
\vskip\baselineskip
\vspace{-0.5cm}

\begin{subfigure}{.08\textwidth}
\centering
  \includegraphics[width=1.\linewidth]{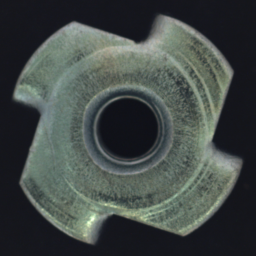}
\end{subfigure}
\begin{subfigure}{.08\textwidth}
\centering
  \includegraphics[width=1.\linewidth]{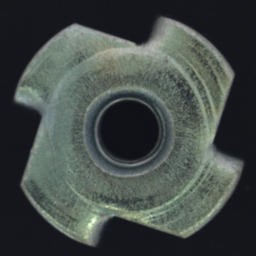}
\end{subfigure}
\begin{subfigure}{.08\textwidth}
\centering
  \includegraphics[width=1.\linewidth]{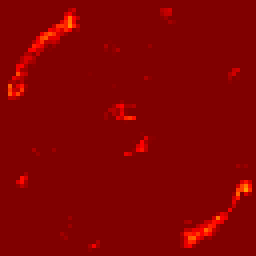}
\end{subfigure}
\begin{subfigure}{.08\textwidth}
\centering
  \includegraphics[width=1.\linewidth]{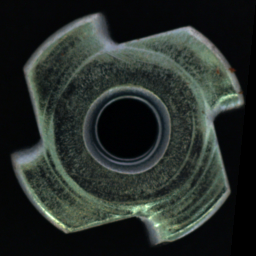}
\end{subfigure}
\begin{subfigure}{.08\textwidth}
\centering
  \includegraphics[width=1.\linewidth]{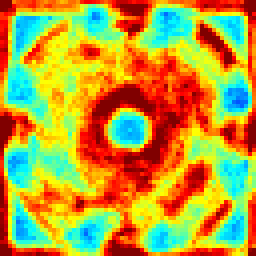}
\end{subfigure}
\begin{subfigure}{.08\textwidth}
\centering
  \includegraphics[width=1.\linewidth]{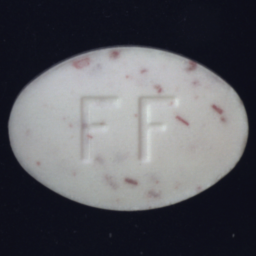}
\end{subfigure}
\begin{subfigure}{.08\textwidth}
\centering
  \includegraphics[width=1.\linewidth]{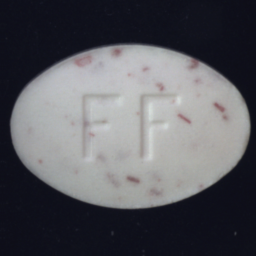}
\end{subfigure}
\begin{subfigure}{.08\textwidth}
\centering
  \includegraphics[width=1.\linewidth]{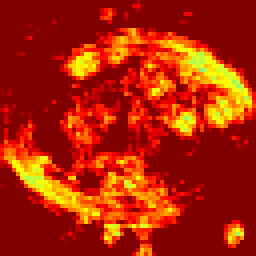}
\end{subfigure}
\begin{subfigure}{.08\textwidth}
\centering
  \includegraphics[width=1.\linewidth]{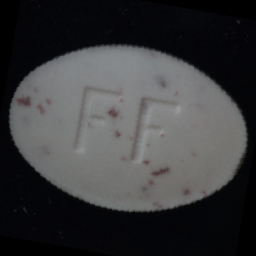}
\end{subfigure}
\begin{subfigure}{.08\textwidth}
\centering
  \includegraphics[width=1.\linewidth]{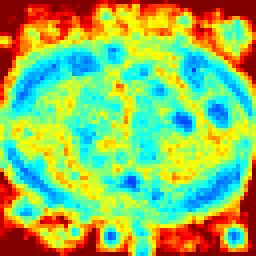}
\end{subfigure}
\vskip\baselineskip
\vspace{-0.5cm}

\begin{subfigure}{.08\textwidth}
\centering
  \includegraphics[width=1.\linewidth]{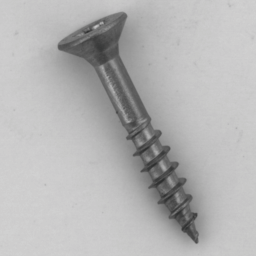}
\end{subfigure}
\begin{subfigure}{.08\textwidth}
\centering
  \includegraphics[width=1.\linewidth]{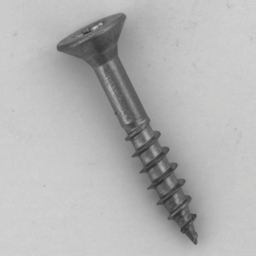}
\end{subfigure}
\begin{subfigure}{.08\textwidth}
\centering
  \includegraphics[width=1.\linewidth]{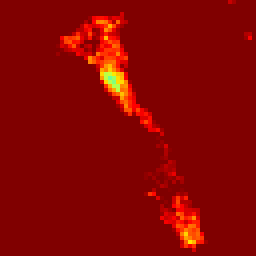}
\end{subfigure}
\begin{subfigure}{.08\textwidth}
\centering
  \includegraphics[width=1.\linewidth]{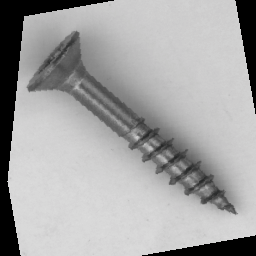}
\end{subfigure}
\begin{subfigure}{.08\textwidth}
\centering
  \includegraphics[width=1.\linewidth]{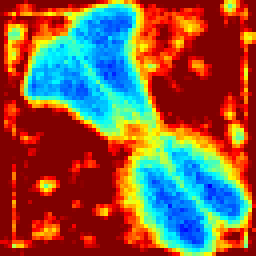}
\end{subfigure}
\begin{subfigure}{.08\textwidth}
\centering
  \includegraphics[width=1.\linewidth]{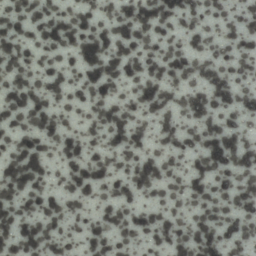}
\end{subfigure}
\begin{subfigure}{.08\textwidth}
\centering
  \includegraphics[width=1.\linewidth]{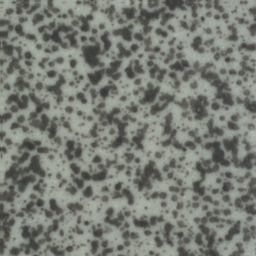}
\end{subfigure}
\begin{subfigure}{.08\textwidth}
\centering
  \includegraphics[width=1.\linewidth]{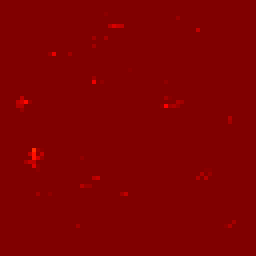}
\end{subfigure}
\begin{subfigure}{.08\textwidth}
\centering
  \includegraphics[width=1.\linewidth]{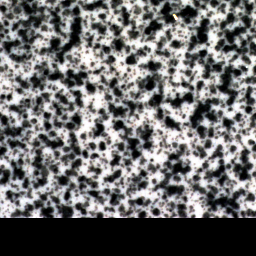}
\end{subfigure}
\begin{subfigure}{.08\textwidth}
\centering
  \includegraphics[width=1.\linewidth]{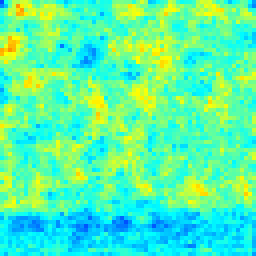}
\end{subfigure}
\vskip\baselineskip
\vspace{-0.5cm}

\begin{subfigure}{.08\textwidth}
\centering
  \includegraphics[width=1.\linewidth]{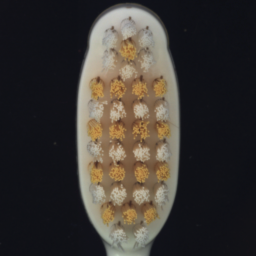}
\end{subfigure}
\begin{subfigure}{.08\textwidth}
\centering
  \includegraphics[width=1.\linewidth]{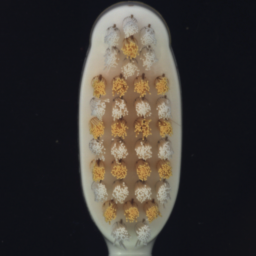}
\end{subfigure}
\begin{subfigure}{.08\textwidth}
\centering
  \includegraphics[width=1.\linewidth]{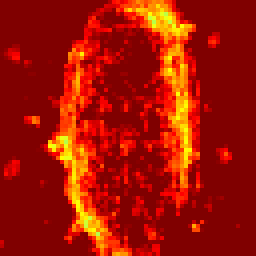}
\end{subfigure}
\begin{subfigure}{.08\textwidth}
\centering
  \includegraphics[width=1.\linewidth]{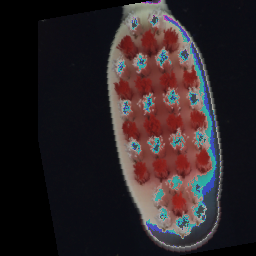}
\end{subfigure}
\begin{subfigure}{.08\textwidth}
\centering
  \includegraphics[width=1.\linewidth]{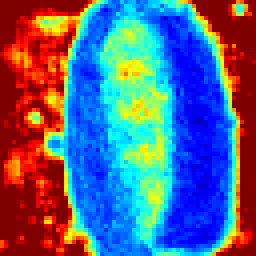}
\end{subfigure}
\begin{subfigure}{.08\textwidth}
\centering
  \includegraphics[width=1.\linewidth]{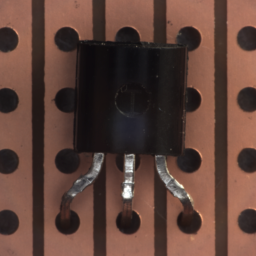}
\end{subfigure}
\begin{subfigure}{.08\textwidth}
\centering
  \includegraphics[width=1.\linewidth]{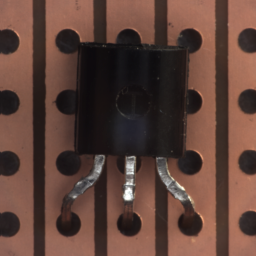}
\end{subfigure}
\begin{subfigure}{.08\textwidth}
\centering
  \includegraphics[width=1.\linewidth]{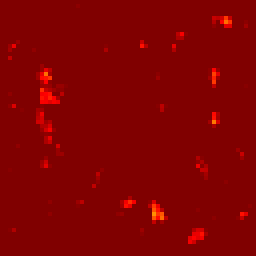}
\end{subfigure}
\begin{subfigure}{.08\textwidth}
\centering
  \includegraphics[width=1.\linewidth]{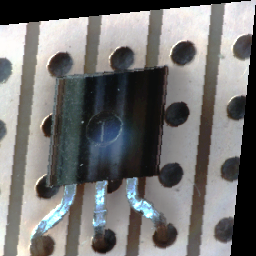}
\end{subfigure}
\begin{subfigure}{.08\textwidth}
\centering
  \includegraphics[width=1.\linewidth]{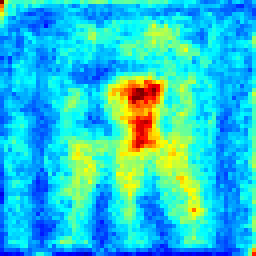}
\end{subfigure}
\vskip\baselineskip
\vspace{-0.5cm}

\begin{subfigure}{.08\textwidth}
\centering
  \includegraphics[width=1.\linewidth]{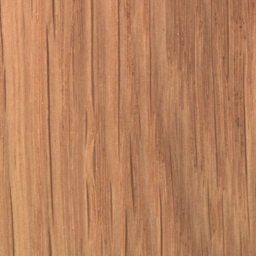}
\end{subfigure}
\begin{subfigure}{.08\textwidth}
\centering
  \includegraphics[width=1.\linewidth]{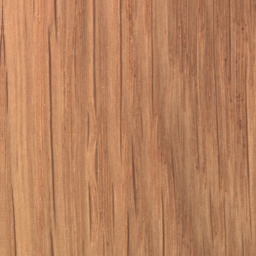}
\end{subfigure}
\begin{subfigure}{.08\textwidth}
\centering
  \includegraphics[width=1.\linewidth]{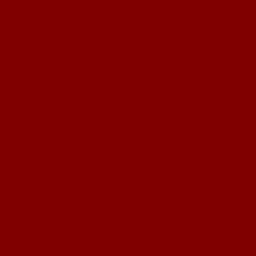}
\end{subfigure}
\begin{subfigure}{.08\textwidth}
\centering
  \includegraphics[width=1.\linewidth]{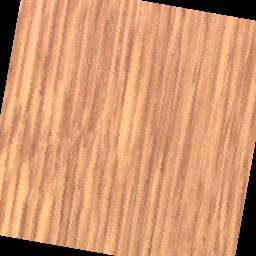}
\end{subfigure}
\begin{subfigure}{.08\textwidth}
\centering
  \includegraphics[width=1.\linewidth]{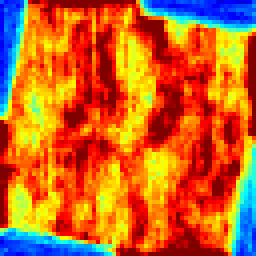}
\end{subfigure}
\begin{subfigure}{.08\textwidth}
\centering
  \includegraphics[width=1.\linewidth]{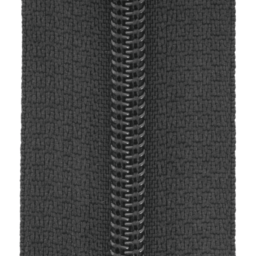}
\end{subfigure}
\begin{subfigure}{.08\textwidth}
\centering
  \includegraphics[width=1.\linewidth]{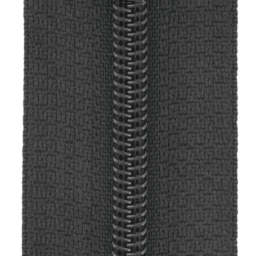}
\end{subfigure}
\begin{subfigure}{.08\textwidth}
\centering
  \includegraphics[width=1.\linewidth]{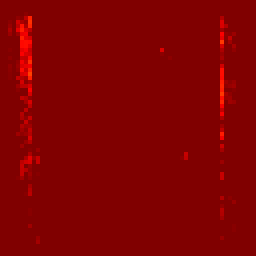}
\end{subfigure}
\begin{subfigure}{.08\textwidth}
\centering
  \includegraphics[width=1.\linewidth]{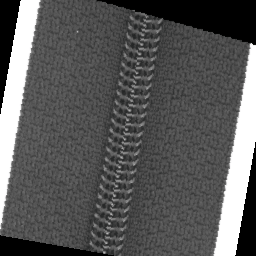}
\end{subfigure}
\begin{subfigure}{.08\textwidth}
\centering
  \includegraphics[width=1.\linewidth]{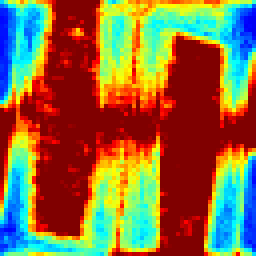}
\end{subfigure}
\vskip\baselineskip
\vspace{-0.5cm}

\begin{subfigure}{.08\textwidth}
\centering
  \includegraphics[width=1.\linewidth]{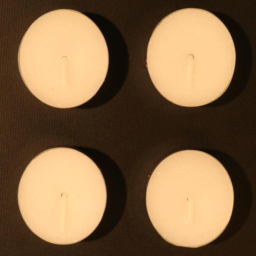}
\end{subfigure}
\begin{subfigure}{.08\textwidth}
\centering
  \includegraphics[width=1.\linewidth]{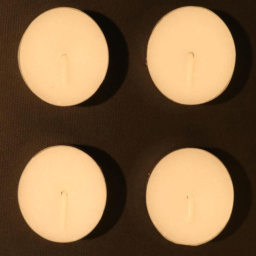}
\end{subfigure}
\begin{subfigure}{.08\textwidth}
\centering
  \includegraphics[width=1.\linewidth]{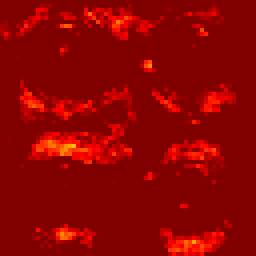}
\end{subfigure}
\begin{subfigure}{.08\textwidth}
\centering
  \includegraphics[width=1.\linewidth]{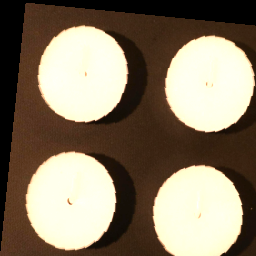}
\end{subfigure}
\begin{subfigure}{.08\textwidth}
\centering
  \includegraphics[width=1.\linewidth]{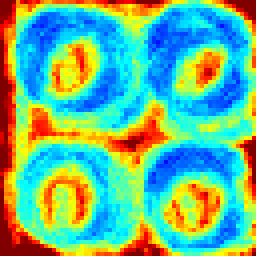}
\end{subfigure}
\begin{subfigure}{.08\textwidth}
\centering
  \includegraphics[width=1.\linewidth]{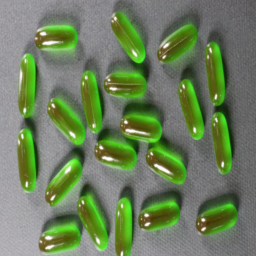}
\end{subfigure}
\begin{subfigure}{.08\textwidth}
\centering
  \includegraphics[width=1.\linewidth]{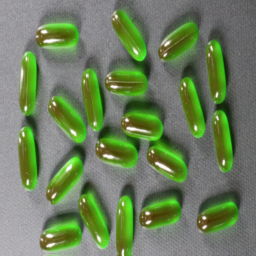}
\end{subfigure}
\begin{subfigure}{.08\textwidth}
\centering
  \includegraphics[width=1.\linewidth]{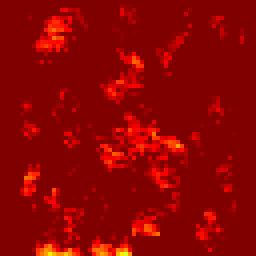}
\end{subfigure}
\begin{subfigure}{.08\textwidth}
\centering
  \includegraphics[width=1.\linewidth]{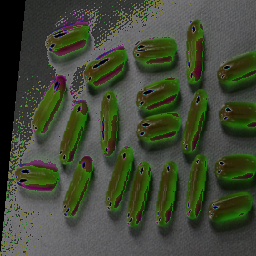}
\end{subfigure}
\begin{subfigure}{.08\textwidth}
\centering
  \includegraphics[width=1.\linewidth]{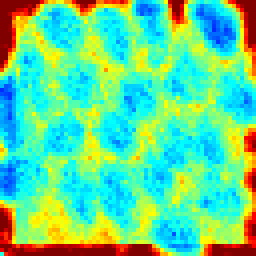}
\end{subfigure}
\vskip\baselineskip
\vspace{-0.5cm}

\begin{subfigure}{.08\textwidth}
\centering
  \includegraphics[width=1.\linewidth]{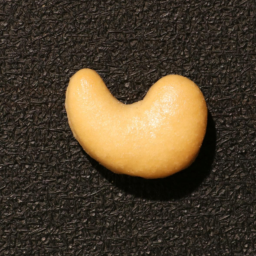}
\end{subfigure}
\begin{subfigure}{.08\textwidth}
\centering
  \includegraphics[width=1.\linewidth]{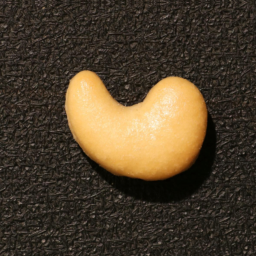}
\end{subfigure}
\begin{subfigure}{.08\textwidth}
\centering
  \includegraphics[width=1.\linewidth]{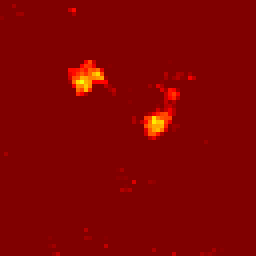}
\end{subfigure}
\begin{subfigure}{.08\textwidth}
\centering
  \includegraphics[width=1.\linewidth]{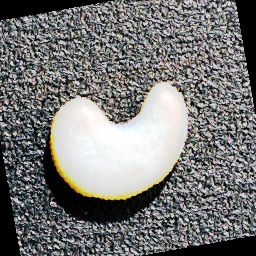}
\end{subfigure}
\begin{subfigure}{.08\textwidth}
\centering
  \includegraphics[width=1.\linewidth]{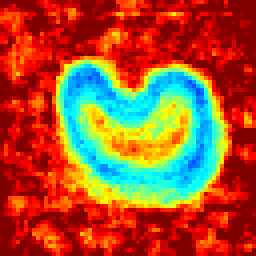}
\end{subfigure}
\begin{subfigure}{.08\textwidth}
\centering
  \includegraphics[width=1.\linewidth]{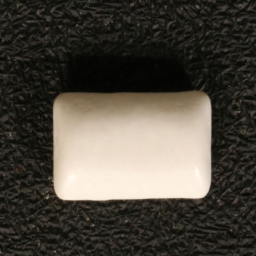}
\end{subfigure}
\begin{subfigure}{.08\textwidth}
\centering
  \includegraphics[width=1.\linewidth]{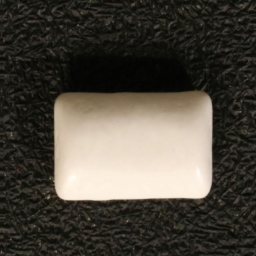}
\end{subfigure}
\begin{subfigure}{.08\textwidth}
\centering
  \includegraphics[width=1.\linewidth]{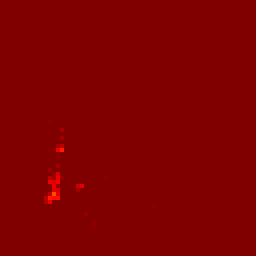}
\end{subfigure}
\begin{subfigure}{.08\textwidth}
\centering
  \includegraphics[width=1.\linewidth]{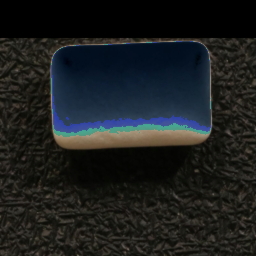}
\end{subfigure}
\begin{subfigure}{.08\textwidth}
\centering
  \includegraphics[width=1.\linewidth]{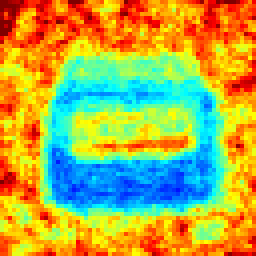}
\end{subfigure}
\vskip\baselineskip
\vspace{-0.5cm}

\begin{subfigure}{.08\textwidth}
\centering
  \includegraphics[width=1.\linewidth]{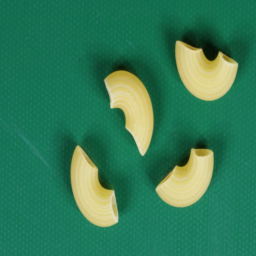}
\end{subfigure}
\begin{subfigure}{.08\textwidth}
\centering
  \includegraphics[width=1.\linewidth]{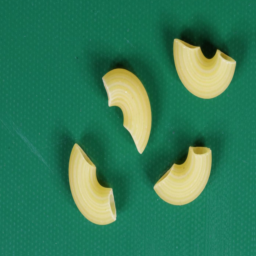}
\end{subfigure}
\begin{subfigure}{.08\textwidth}
\centering
  \includegraphics[width=1.\linewidth]{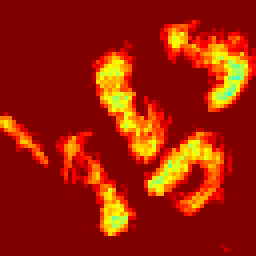}
\end{subfigure}
\begin{subfigure}{.08\textwidth}
\centering
  \includegraphics[width=1.\linewidth]{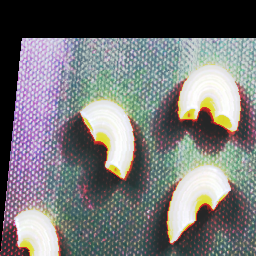}
\end{subfigure}
\begin{subfigure}{.08\textwidth}
\centering
  \includegraphics[width=1.\linewidth]{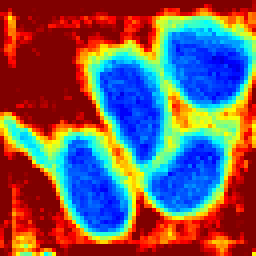}
\end{subfigure}
\begin{subfigure}{.08\textwidth}
\centering
  \includegraphics[width=1.\linewidth]{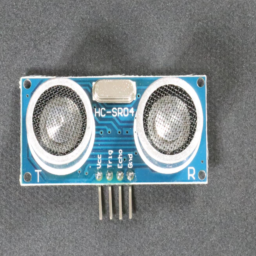}
\end{subfigure}
\begin{subfigure}{.08\textwidth}
\centering
  \includegraphics[width=1.\linewidth]{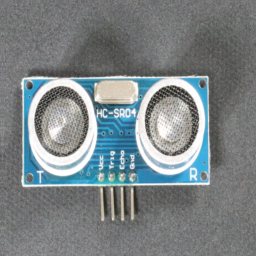}
\end{subfigure}
\begin{subfigure}{.08\textwidth}
\centering
  \includegraphics[width=1.\linewidth]{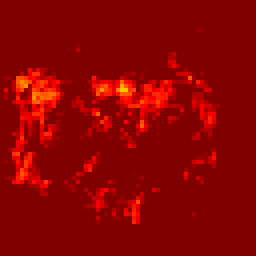}
\end{subfigure}
\begin{subfigure}{.08\textwidth}
\centering
  \includegraphics[width=1.\linewidth]{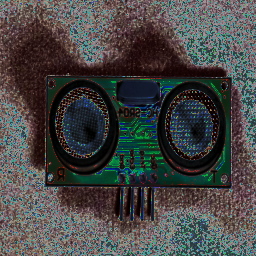}
\end{subfigure}
\begin{subfigure}{.08\textwidth}
\centering
  \includegraphics[width=1.\linewidth]{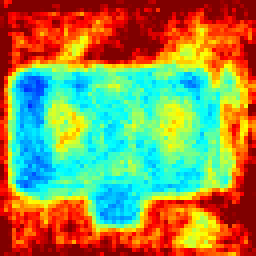}
 \end{subfigure}
\vskip\baselineskip
\vspace{-0.5cm}

\begin{subfigure}{.08\textwidth}
\centering
  \includegraphics[width=1.\linewidth]{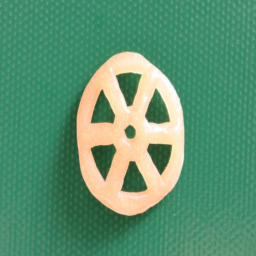}
\end{subfigure}
\begin{subfigure}{.08\textwidth}
\centering
  \includegraphics[width=1.\linewidth]{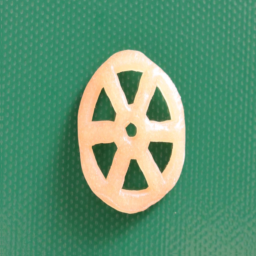}
\end{subfigure}
\begin{subfigure}{.08\textwidth}
\centering
  \includegraphics[width=1.\linewidth]{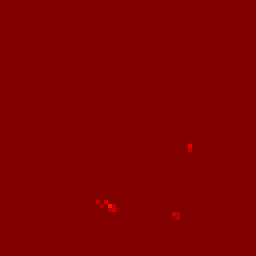}
\end{subfigure}
\begin{subfigure}{.08\textwidth}
\centering
  \includegraphics[width=1.\linewidth]{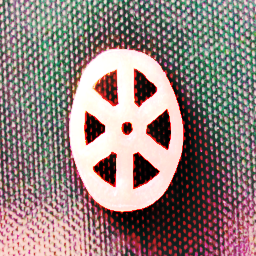}
\end{subfigure}
\begin{subfigure}{.08\textwidth}
\centering
  \includegraphics[width=1.\linewidth]{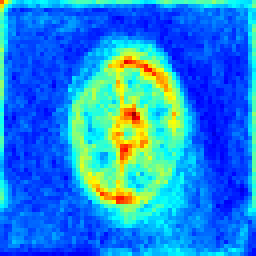}
\end{subfigure}
\begin{subfigure}{.08\textwidth}
\centering
  \includegraphics[width=1.\linewidth]{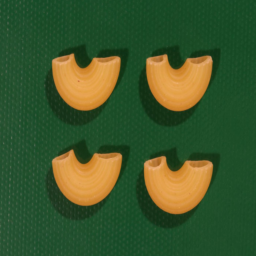}
\end{subfigure}
\begin{subfigure}{.08\textwidth}
\centering
  \includegraphics[width=1.\linewidth]{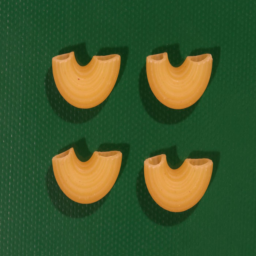}
\end{subfigure}
\begin{subfigure}{.08\textwidth}
\centering
  \includegraphics[width=1.\linewidth]{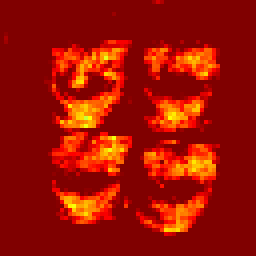}
\end{subfigure}
\begin{subfigure}{.08\textwidth}
\centering
  \includegraphics[width=1.\linewidth]{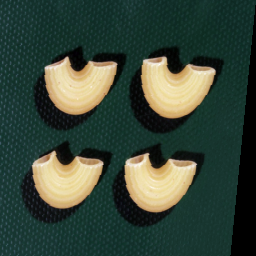}
\end{subfigure}
\begin{subfigure}{.08\textwidth}
\centering
  \includegraphics[width=1.\linewidth]{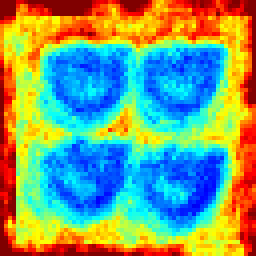}
\end{subfigure}
\vskip\baselineskip
\vspace{-0.5cm}

\begin{subfigure}{.08\textwidth}
\centering
  \includegraphics[width=1.\linewidth]{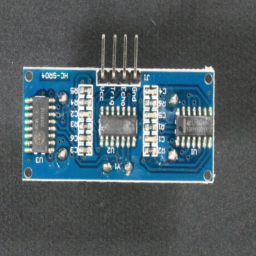}
\end{subfigure}
\begin{subfigure}{.08\textwidth}
\centering
  \includegraphics[width=1.\linewidth]{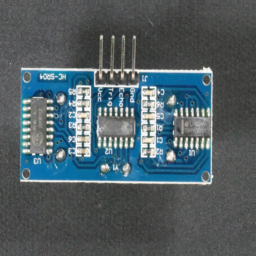}
\end{subfigure}
\begin{subfigure}{.08\textwidth}
\centering
  \includegraphics[width=1.\linewidth]{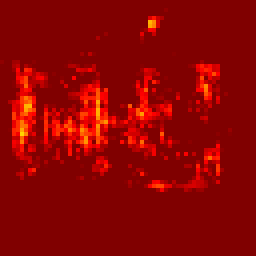}
\end{subfigure}
\begin{subfigure}{.08\textwidth}
\centering
  \includegraphics[width=1.\linewidth]{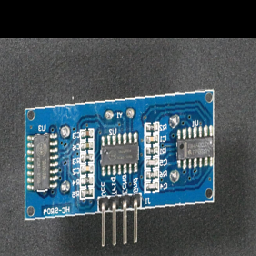}
\end{subfigure}
\begin{subfigure}{.08\textwidth}
\centering
  \includegraphics[width=1.\linewidth]{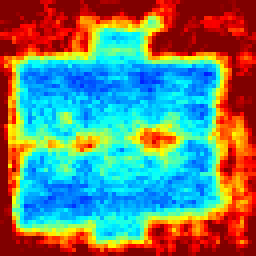}
\end{subfigure}
\begin{subfigure}{.08\textwidth}
\centering
  \includegraphics[width=1.\linewidth]{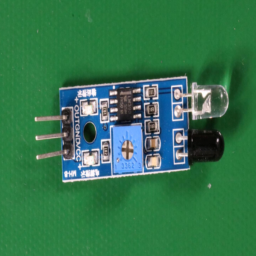}
\end{subfigure}
\begin{subfigure}{.08\textwidth}
\centering
  \includegraphics[width=1.\linewidth]{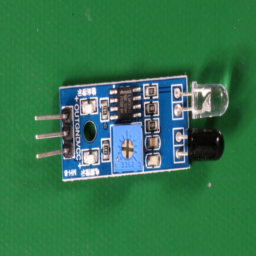}
\end{subfigure}
\begin{subfigure}{.08\textwidth}
\centering
  \includegraphics[width=1.\linewidth]{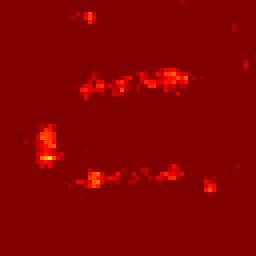}
\end{subfigure}
\begin{subfigure}{.08\textwidth}
\centering
  \includegraphics[width=1.\linewidth]{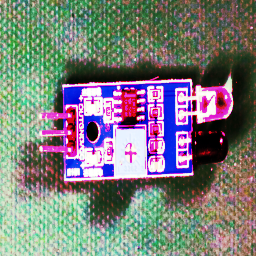}
\end{subfigure}
\begin{subfigure}{.08\textwidth}
\centering
  \includegraphics[width=1.\linewidth]{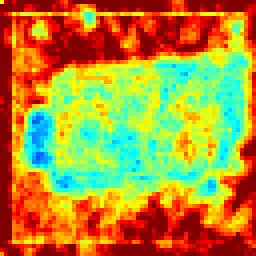}
 \end{subfigure}
\vskip\baselineskip
\vspace{-0.5cm}

\begin{subfigure}{.08\textwidth}
\centering
  \includegraphics[width=1.\linewidth]{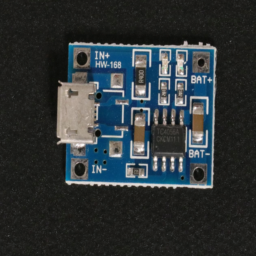}
  \fontsize{0.25cm}{0.25cm}\selectfont{{(a)}}
\end{subfigure}
\begin{subfigure}{.08\textwidth}
\centering
  \includegraphics[width=1.\linewidth]{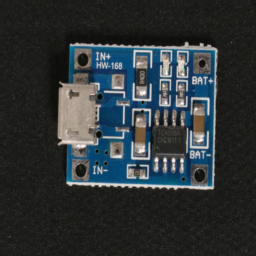}
  \fontsize{0.25cm}{0.25cm}\selectfont{{(b)}}
\end{subfigure}
\begin{subfigure}{.08\textwidth}
\centering
  \includegraphics[width=1.\linewidth]{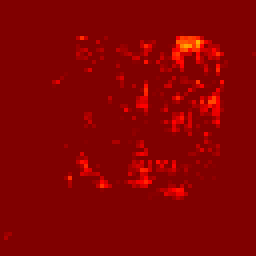}
  \fontsize{0.25cm}{0.25cm}\selectfont{{(c)}}
\end{subfigure}
\begin{subfigure}{.08\textwidth}
\centering
  \includegraphics[width=1.\linewidth]{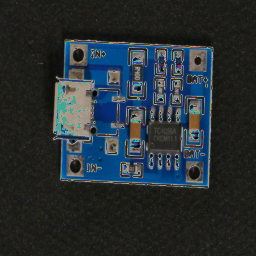}
  \fontsize{0.25cm}{0.25cm}\selectfont{{(d)}}
\end{subfigure}
\begin{subfigure}{.08\textwidth}
\centering
  \includegraphics[width=1.\linewidth]{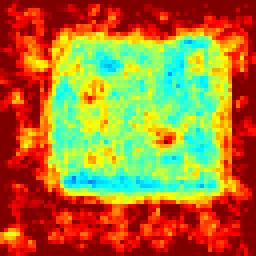}
  \fontsize{0.25cm}{0.25cm}\selectfont{{(e)}}
\end{subfigure}
\begin{subfigure}{.08\textwidth}
\centering
  \includegraphics[width=1.\linewidth]{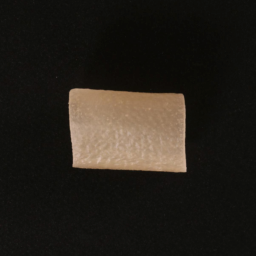}
  \fontsize{0.25cm}{0.25cm}\selectfont{{(a)}}
\end{subfigure}
\begin{subfigure}{.08\textwidth}
\centering
  \includegraphics[width=1.\linewidth]{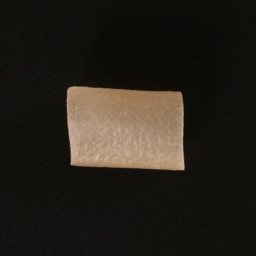}
  \fontsize{0.25cm}{0.25cm}\selectfont{{(b)}}
\end{subfigure}
\begin{subfigure}{.08\textwidth}
\centering
  \includegraphics[width=1.\linewidth]{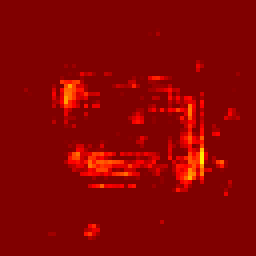}
  \fontsize{0.25cm}{0.25cm}\selectfont{{(c)}}
\end{subfigure}
\begin{subfigure}{.08\textwidth}
\centering
  \includegraphics[width=1.\linewidth]{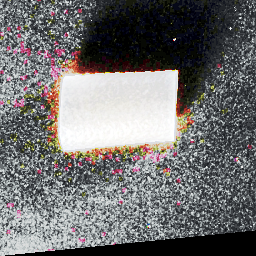}
  \fontsize{0.25cm}{0.25cm}\selectfont{{(d)}}
\end{subfigure}
\begin{subfigure}{.08\textwidth}
\centering
  \includegraphics[width=1.\linewidth]{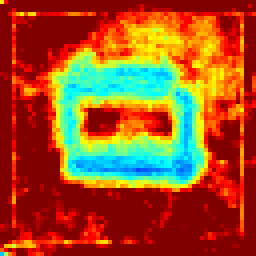}
  \fontsize{0.25cm}{0.25cm}\selectfont{{(e)}}
\end{subfigure}
\vskip\baselineskip
\vspace{-0.5cm}
\end{center}
\vspace{-0.5cm}
\caption{{The region for updating features in augmented images.} The columns from left to right represent (a) original images, (b) weak augmented images, (c) the location and intensity region used in feature learning for the weakly augmentation, (d) a strongly augmented image, and (e) the location and intensity region for the strong augmentation.
In (c) and (e), the color scale signifies that a shift toward red corresponds to a larger number of channels employed for learning in the feature, while a shift toward blue indicates the use of fewer channels.}
\label{fig:ablationA}
\end{figure*}

\begin{figure*}[t]
    \centering
    \small
    \resizebox{1.\linewidth}{!}{
    \setlength{\tabcolsep}{.7pt}
    \begin{tabular}{cccccccccccc}
        {\includegraphics[width=1.\linewidth]{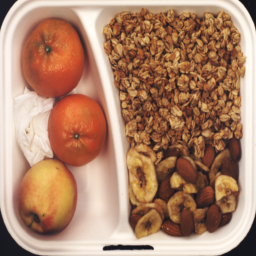}} &{\includegraphics[width=1.\linewidth]{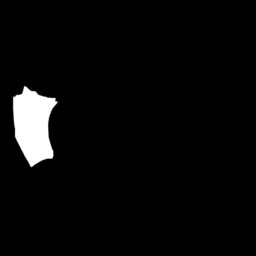}} &{\includegraphics[width=1.\linewidth]{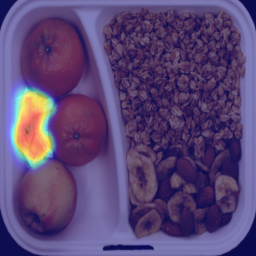}} & 
        \hspace{2.5cm}
        {\includegraphics[width=1.\linewidth]{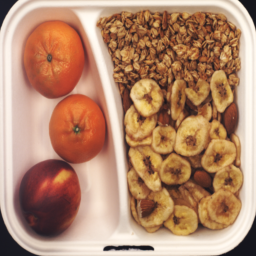}} &{\includegraphics[width=1.\linewidth]{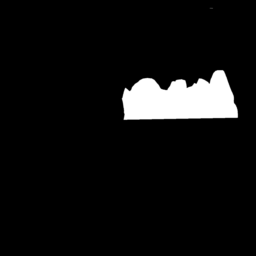}} &{\includegraphics[width=1.\linewidth]{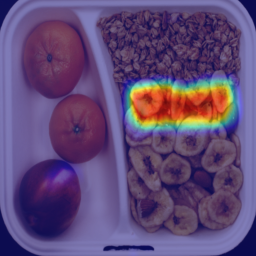}} & 
        \hspace{2.5cm}
        {\includegraphics[width=1.\linewidth]{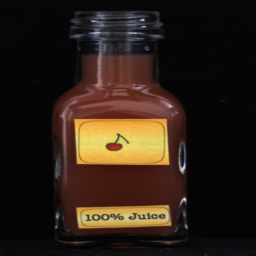}} &{\includegraphics[width=1.\linewidth]{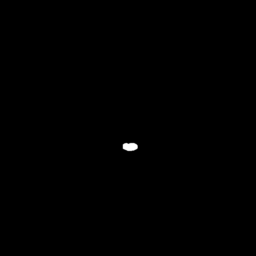}} &{\includegraphics[width=1.\linewidth]{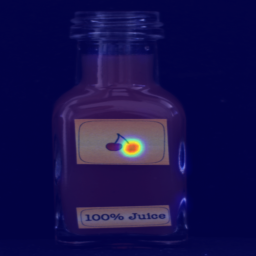}} & 
        \hspace{2.5cm}
        {\includegraphics[width=1.\linewidth]{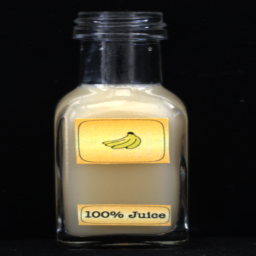}} &{\includegraphics[width=1.\linewidth]{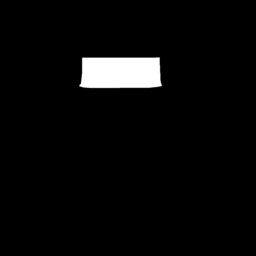}} & {\includegraphics[width=1.\linewidth]{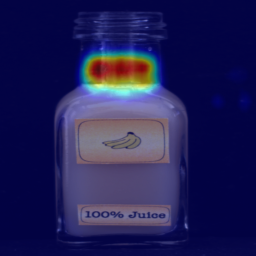}}\\
    
        {\includegraphics[width=1.\linewidth]{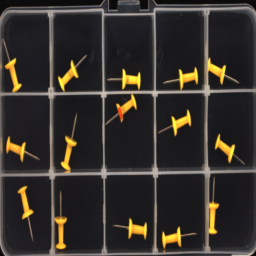}} &{\includegraphics[width=1.\linewidth]{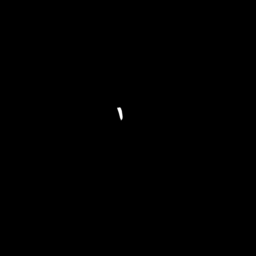}} &{\includegraphics[width=1.\linewidth]{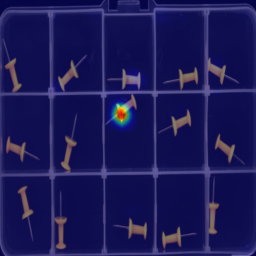}} & 
        \hspace{2.5cm}
        {\includegraphics[width=1.\linewidth]{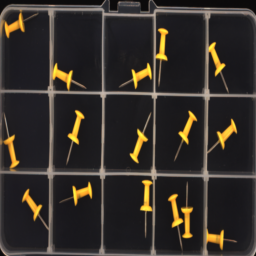}} &{\includegraphics[width=1.\linewidth]{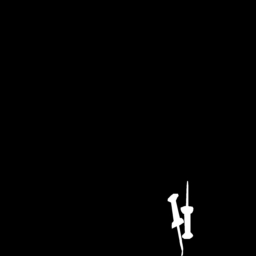}} &{\includegraphics[width=1.\linewidth]{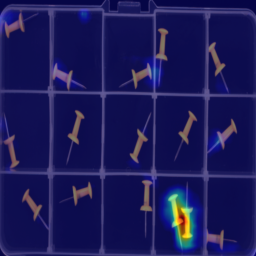}} & 
        \hspace{2.5cm}
        {\includegraphics[width=1.\linewidth]{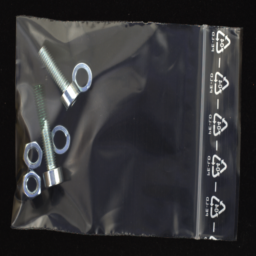}} &{\includegraphics[width=1.\linewidth]{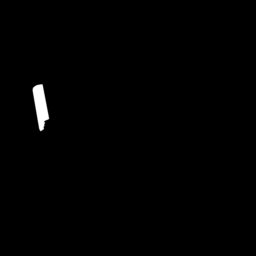}} &{\includegraphics[width=1.\linewidth]{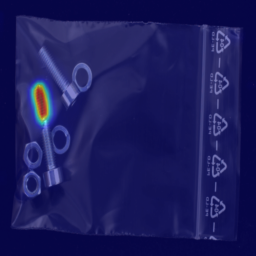}} & 
        \hspace{2.5cm}
        {\includegraphics[width=1.\linewidth]{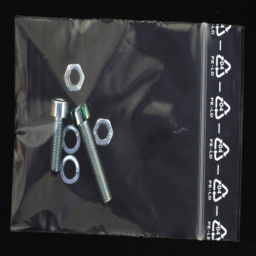}} &{\includegraphics[width=1.\linewidth]{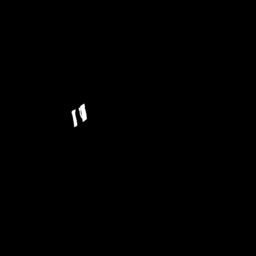}} & {\includegraphics[width=1.\linewidth]{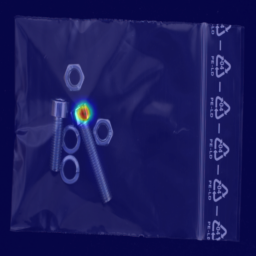}}\\

        {\includegraphics[width=1.\linewidth]{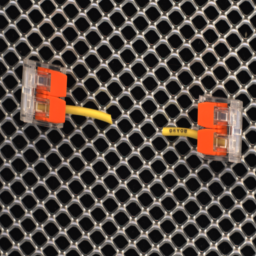}} &{\includegraphics[width=1.\linewidth]{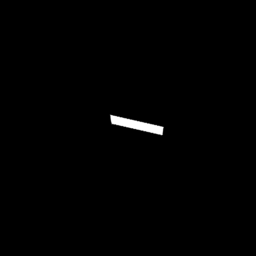}} &{\includegraphics[width=1.\linewidth]{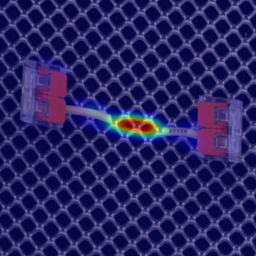}} & 
        \hspace{2.5cm}
        {\includegraphics[width=1.\linewidth]{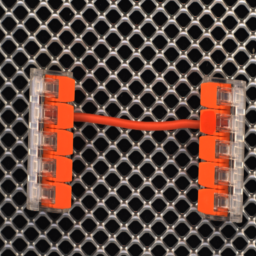}} &{\includegraphics[width=1.\linewidth]{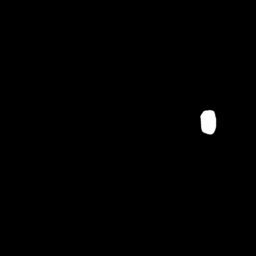}} &{\includegraphics[width=1.\linewidth]{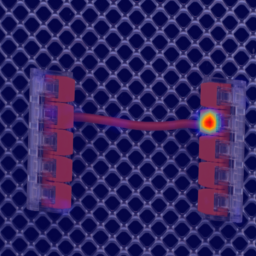}} & 
        \hspace{2.5cm}
        {\includegraphics[width=1.\linewidth]{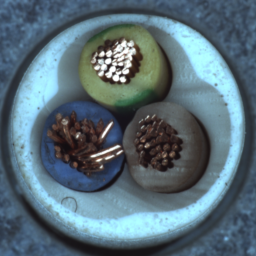}} &{\includegraphics[width=1.\linewidth]{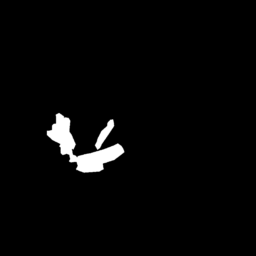}} &{\includegraphics[width=1.\linewidth]{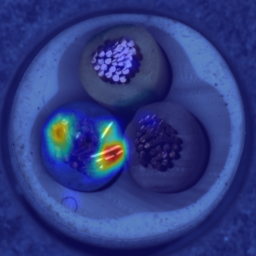}} & 
        \hspace{2.5cm}
        {\includegraphics[width=1.\linewidth]{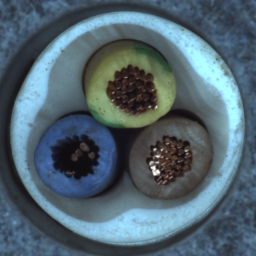}} &{\includegraphics[width=1.\linewidth]{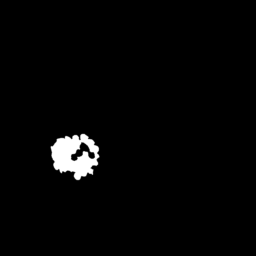}} & {\includegraphics[width=1.\linewidth]{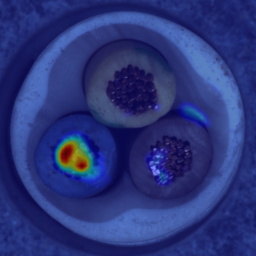}}\\
        
        {\includegraphics[width=1.\linewidth]{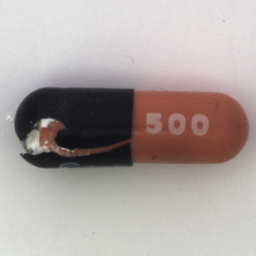}} &{\includegraphics[width=1.\linewidth]{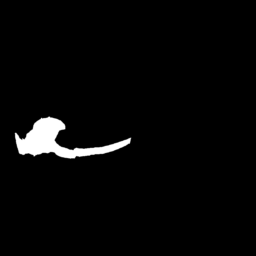}} &{\includegraphics[width=1.\linewidth]{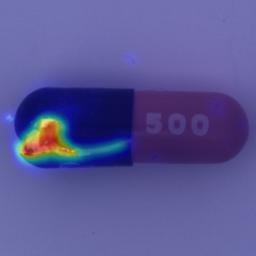}} & 
        \hspace{2.5cm}
        {\includegraphics[width=1.\linewidth]{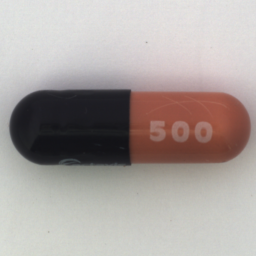}} &{\includegraphics[width=1.\linewidth]{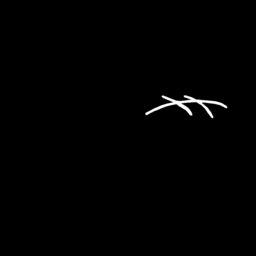}} &    {\includegraphics[width=1.\linewidth]{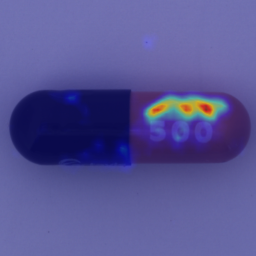}} &
        \hspace{2.5cm}
        {\includegraphics[width=1.\linewidth]{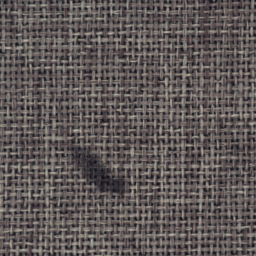}} &{\includegraphics[width=1.\linewidth]{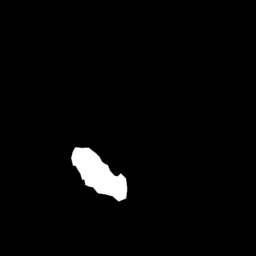}} &{\includegraphics[width=1.\linewidth]{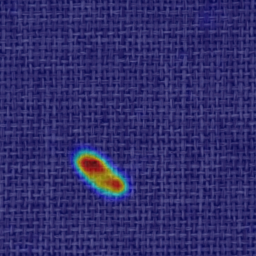}} & 
        \hspace{2.5cm}
        {\includegraphics[width=1.\linewidth]{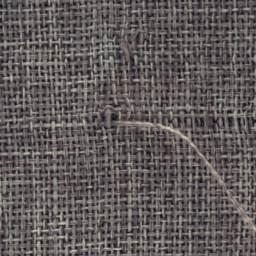}} &{\includegraphics[width=1.\linewidth]{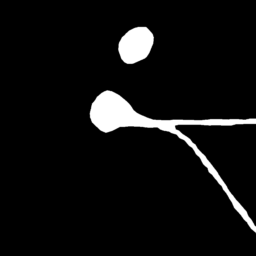}} &         {\includegraphics[width=1.\linewidth]{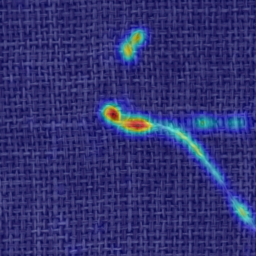}}\\
        
        {\includegraphics[width=1.\linewidth]{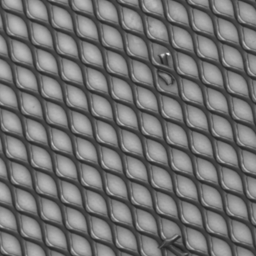}} &{\includegraphics[width=1.\linewidth]{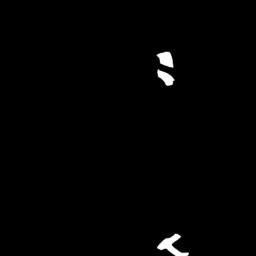}} &{\includegraphics[width=1.\linewidth]{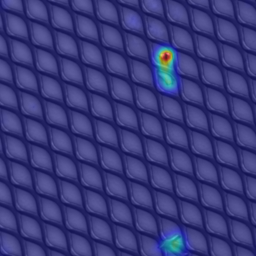}} &  
        \hspace{2.5cm}
        {\includegraphics[width=1.\linewidth]{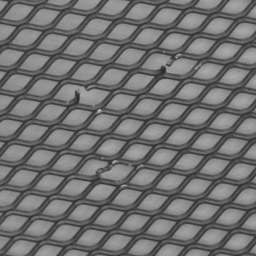}} &{\includegraphics[width=1.\linewidth]{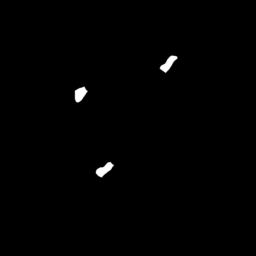}} &         {\includegraphics[width=1.\linewidth]{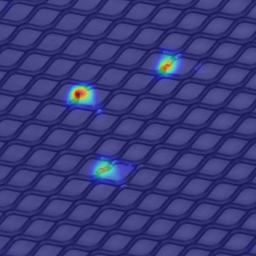}} &
        \hspace{2.5cm}
        {\includegraphics[width=1.\linewidth]{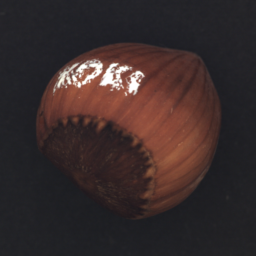}} &{\includegraphics[width=1.\linewidth]{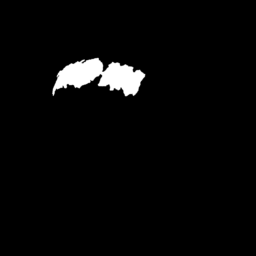}} &{\includegraphics[width=1.\linewidth]{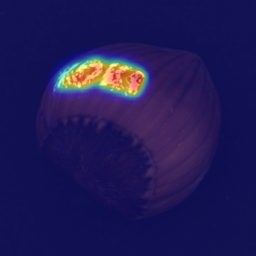}} &
        \hspace{2.5cm}
        {\includegraphics[width=1.\linewidth]{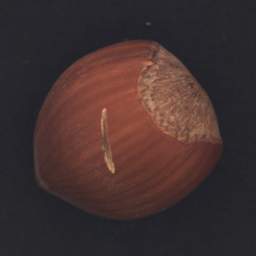}} &{\includegraphics[width=1.\linewidth]{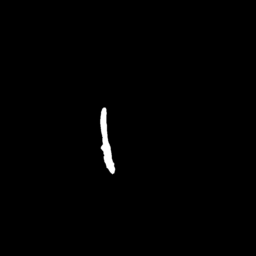}} &     {\includegraphics[width=1.\linewidth]{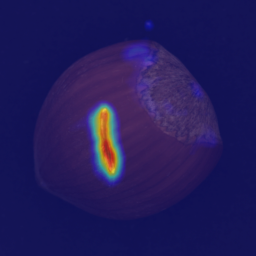}}\\

        {\includegraphics[width=1.\linewidth]{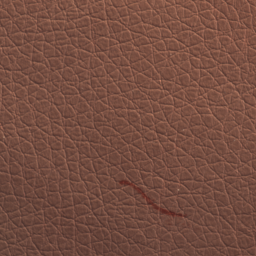}} &{\includegraphics[width=1.\linewidth]{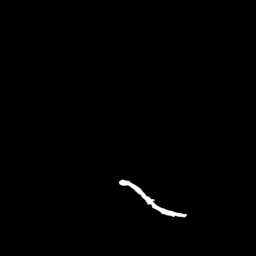}} &{\includegraphics[width=1.\linewidth]{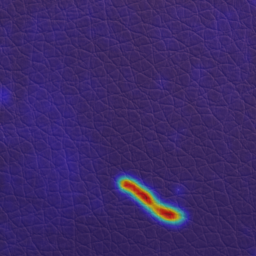}} &
        \hspace{2.5cm}
        {\includegraphics[width=1.\linewidth]{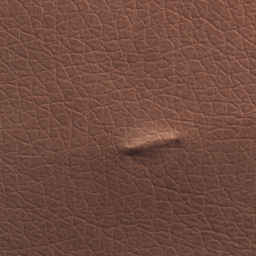}} &{\includegraphics[width=1.\linewidth]{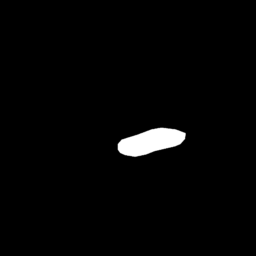}} &      {\includegraphics[width=1.\linewidth]{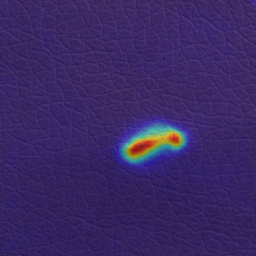}} &
        \hspace{2.5cm}
        {\includegraphics[width=1.\linewidth]{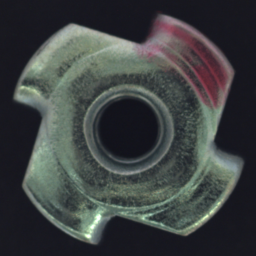}} &{\includegraphics[width=1.\linewidth]{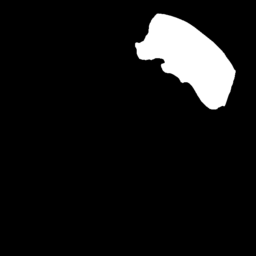}} &{\includegraphics[width=1.\linewidth]{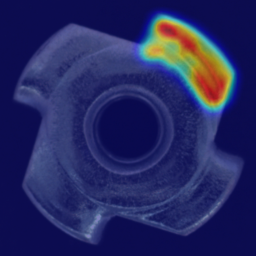}} &
        \hspace{2.5cm}
        {\includegraphics[width=1.\linewidth]{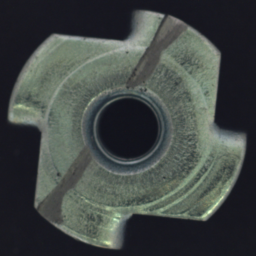}} &{\includegraphics[width=1.\linewidth]{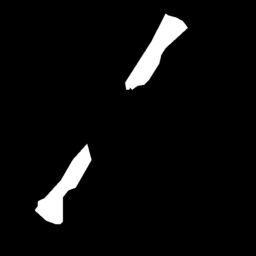}} &     {\includegraphics[width=1.\linewidth]{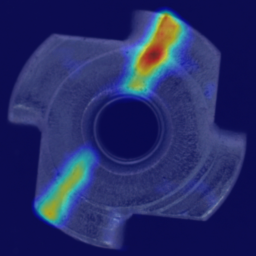}}\\

        {\includegraphics[width=1.\linewidth]{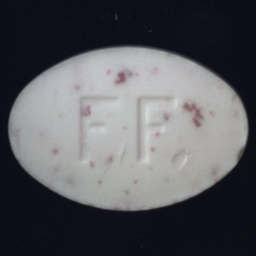}} &{\includegraphics[width=1.\linewidth]{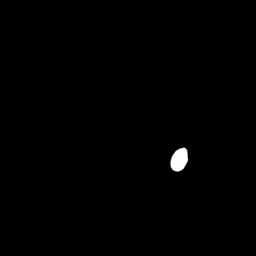}} &{\includegraphics[width=1.\linewidth]{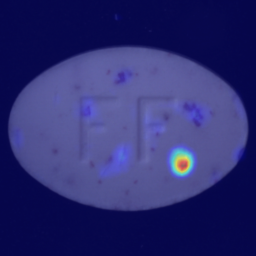}} &  
        \hspace{2.5cm}
        {\includegraphics[width=1.\linewidth]{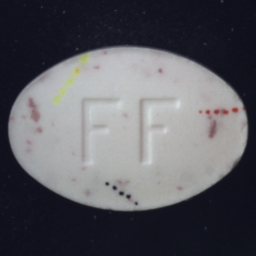}} &{\includegraphics[width=1.\linewidth]{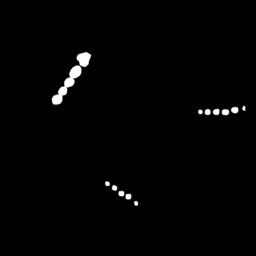}} &         {\includegraphics[width=1.\linewidth]{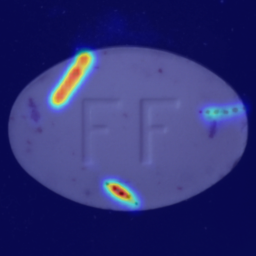}}&
        \hspace{2.5cm}
        {\includegraphics[width=1.\linewidth]{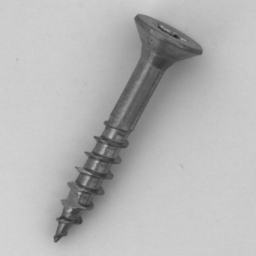}} &{\includegraphics[width=1.\linewidth]{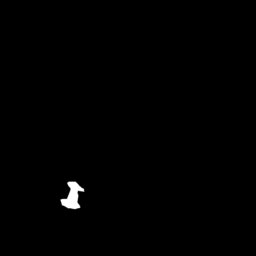}} &{\includegraphics[width=1.\linewidth]{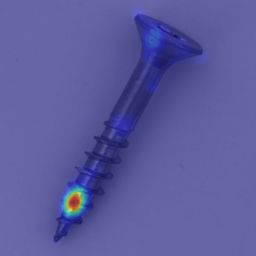}} & 
        \hspace{2.5cm}
        {\includegraphics[width=1.\linewidth]{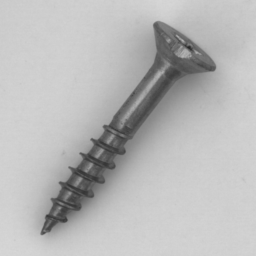}} &{\includegraphics[width=1.\linewidth]{fig/suppl/anomalymap/mvtec/screw/0041.png}} &{\includegraphics[width=1.\linewidth]{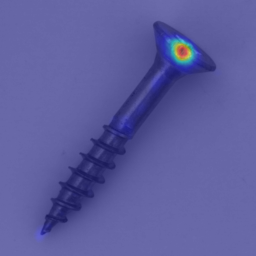}}\\
        
        {\includegraphics[width=1.\linewidth]{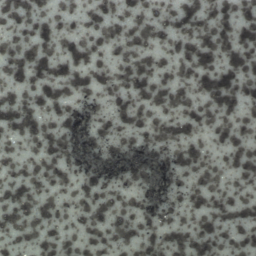}} &{\includegraphics[width=1.\linewidth]{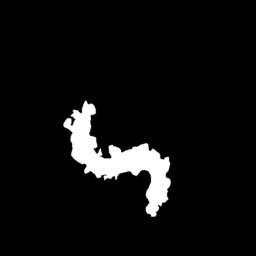}} &{\includegraphics[width=1.\linewidth]{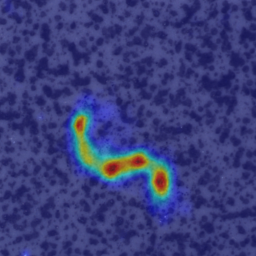}} & 
        \hspace{2.5cm}
        {\includegraphics[width=1.\linewidth]{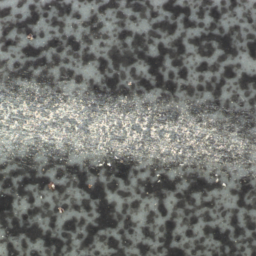}} &{\includegraphics[width=1.\linewidth]{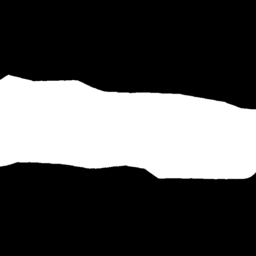}} &         {\includegraphics[width=1.\linewidth]{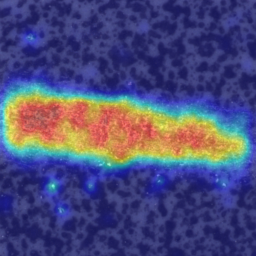}} &
        \hspace{2.5cm}
        {\includegraphics[width=1.\linewidth]{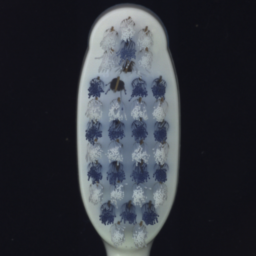}} &{\includegraphics[width=1.\linewidth]{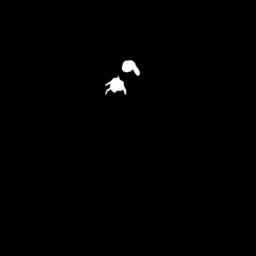}} &{\includegraphics[width=1.\linewidth]{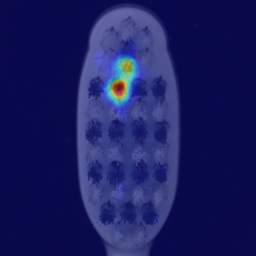}} &
        \hspace{2.5cm}
        {\includegraphics[width=1.\linewidth]{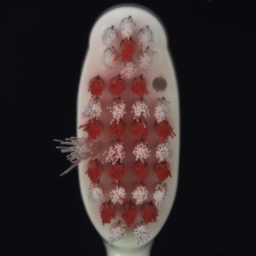}} &{\includegraphics[width=1.\linewidth]{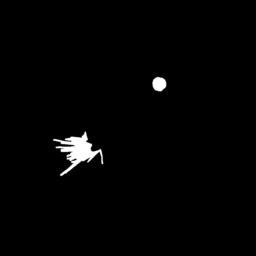}} &       {\includegraphics[width=1.\linewidth]{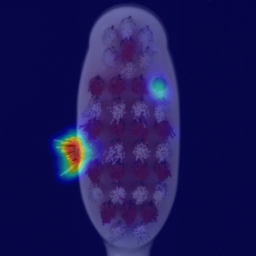}}\\

        {\includegraphics[width=1.\linewidth]{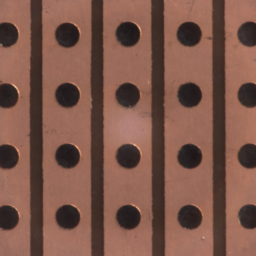}} &{\includegraphics[width=1.\linewidth]{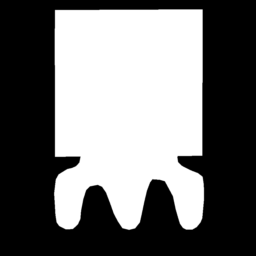}} &{\includegraphics[width=1.\linewidth]{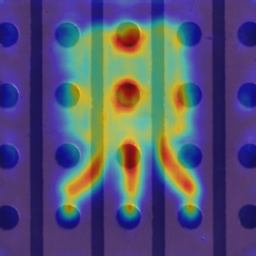}} &
        \hspace{2.5cm}
        {\includegraphics[width=1.\linewidth]{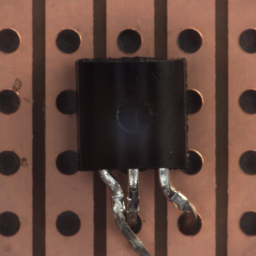}} &{\includegraphics[width=1.\linewidth]{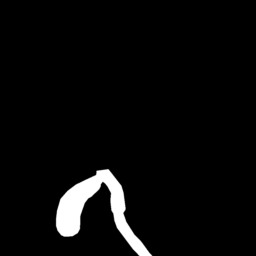}} &       {\includegraphics[width=1.\linewidth]{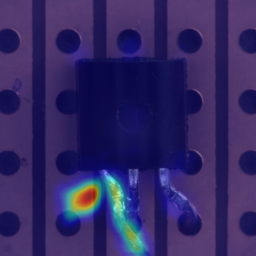}} &
        \hspace{2.5cm}
        {\includegraphics[width=1.\linewidth]{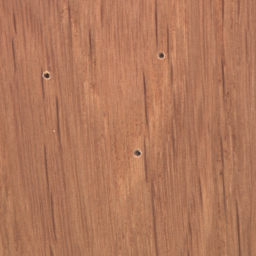}} &{\includegraphics[width=1.\linewidth]{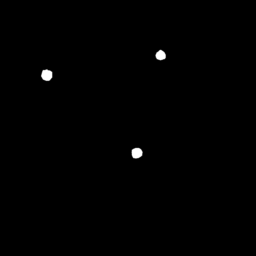}} &{\includegraphics[width=1.\linewidth]{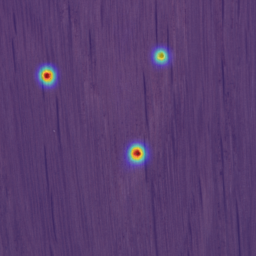}} &  
        \hspace{2.5cm}
        {\includegraphics[width=1.\linewidth]{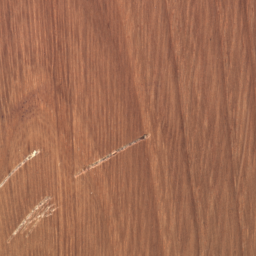}} &{\includegraphics[width=1.\linewidth]{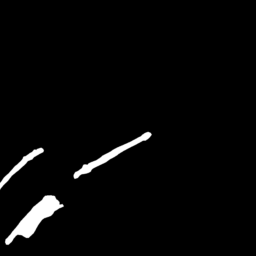}} &         {\includegraphics[width=1.\linewidth]{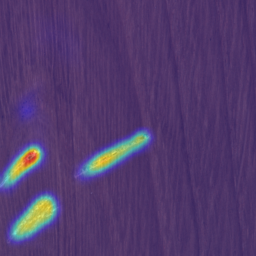}}\\

        {\includegraphics[width=1.\linewidth]{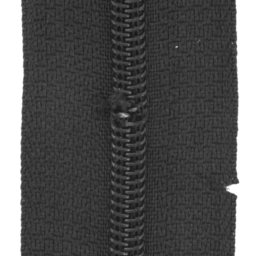}} &{\includegraphics[width=1.\linewidth]{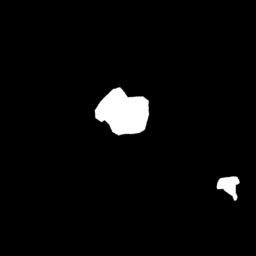}} &{\includegraphics[width=1.\linewidth]{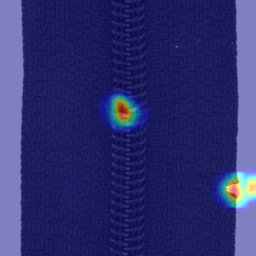}} &  
        \hspace{2.5cm}
        {\includegraphics[width=1.\linewidth]{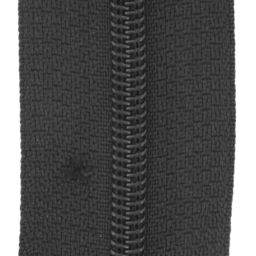}} &{\includegraphics[width=1.\linewidth]{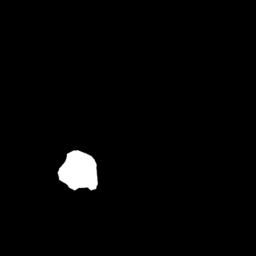}} &         {\includegraphics[width=1.\linewidth]{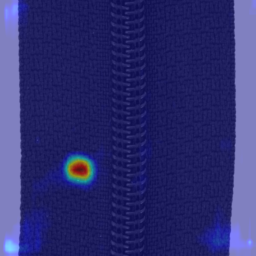}} & 
        \hspace{2.5cm}
        {\includegraphics[width=1.\linewidth]{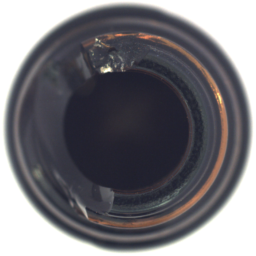}} &{\includegraphics[width=1.\linewidth]{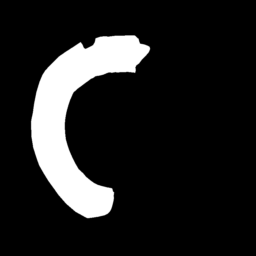}} &{\includegraphics[width=1.\linewidth]{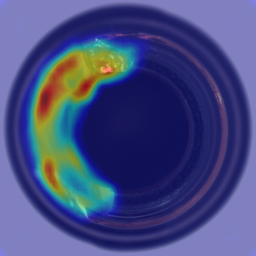}} & 
        \hspace{2.5cm}
        {\includegraphics[width=1.\linewidth]{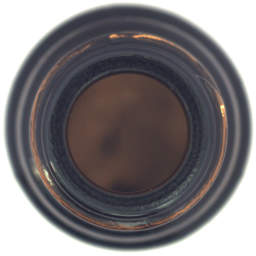}} &{\includegraphics[width=1.\linewidth]{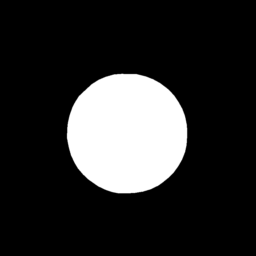}} &{\includegraphics[width=1.\linewidth]{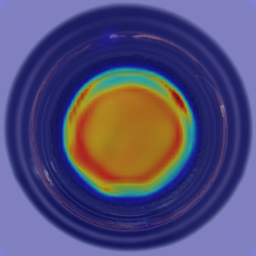}}\\ 
        
        {\includegraphics[width=1.\linewidth]{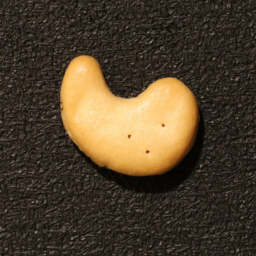}} &{\includegraphics[width=1.\linewidth]{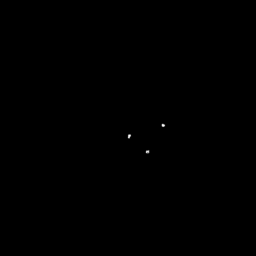}} &{\includegraphics[width=1.\linewidth]{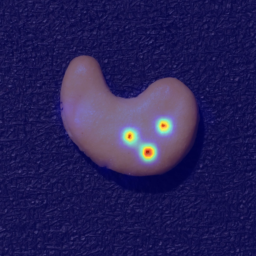}} & 
        \hspace{2.5cm}
        {\includegraphics[width=1.\linewidth]{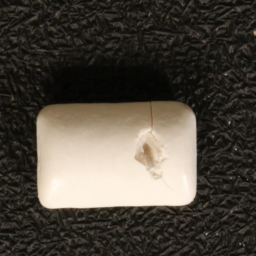}} &{\includegraphics[width=1.\linewidth]{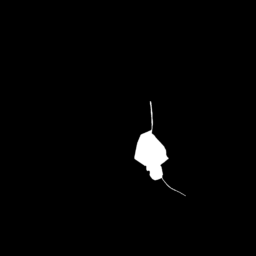}} &        {\includegraphics[width=1.\linewidth]{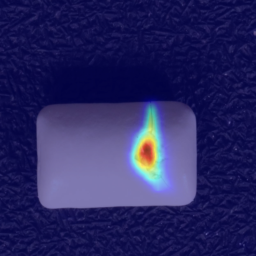}} & 
        \hspace{2.5cm}
        {\includegraphics[width=1.\linewidth]{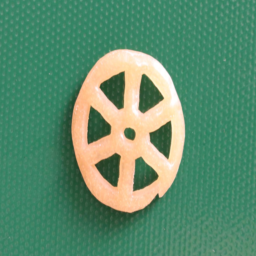}} &{\includegraphics[width=1.\linewidth]{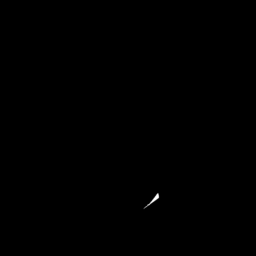}} &{\includegraphics[width=1.\linewidth]{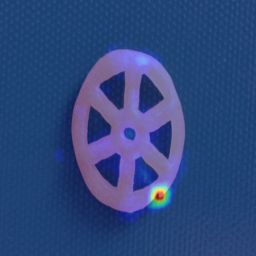}} & 
        \hspace{2.5cm}
        {\includegraphics[width=1.\linewidth]{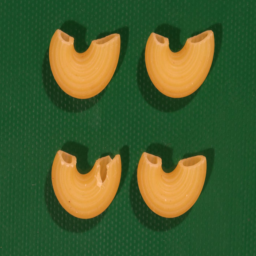}} &{\includegraphics[width=1.\linewidth]{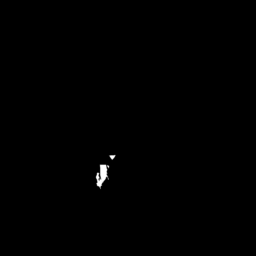}} &         {\includegraphics[width=1.\linewidth]{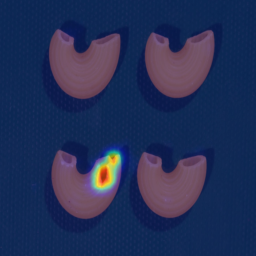}}\\ 

        {\includegraphics[width=1.\linewidth]{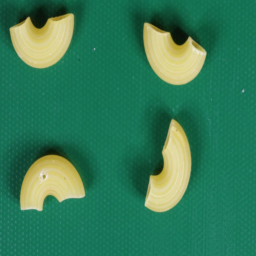}} &{\includegraphics[width=1.\linewidth]{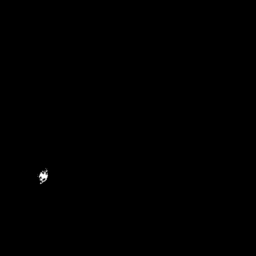}} &{\includegraphics[width=1.\linewidth]{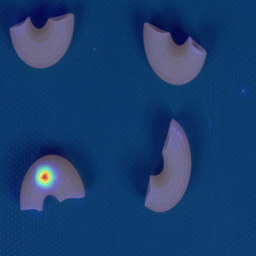}} &
        \hspace{2.5cm}
        {\includegraphics[width=1.\linewidth]{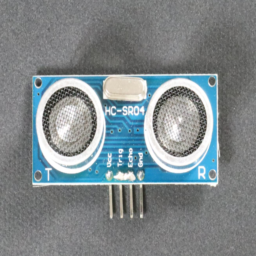}} &
        {\includegraphics[width=1.\linewidth]{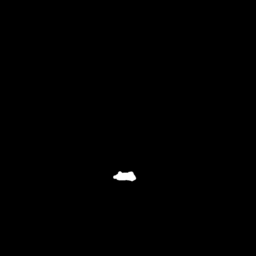}} &         {\includegraphics[width=1.\linewidth]{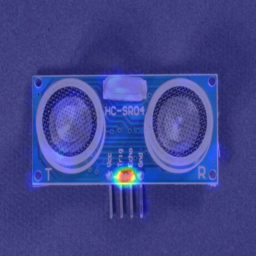}} & 
        \hspace{2.5cm}
        {\includegraphics[width=1.\linewidth]{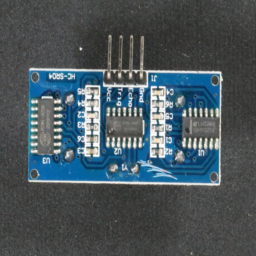}} &
        {\includegraphics[width=1.\linewidth]{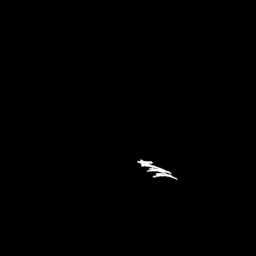}} &
        {\includegraphics[width=1.\linewidth]{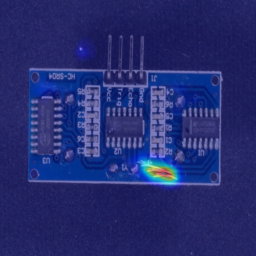}} &  
        \hspace{2.5cm}
        {\includegraphics[width=1.\linewidth]{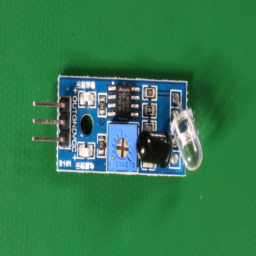}} &
        {\includegraphics[width=1.\linewidth]{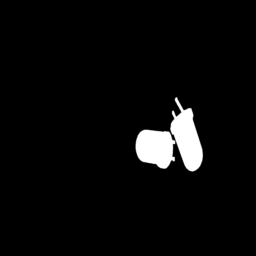}} &         {\includegraphics[width=1.\linewidth]{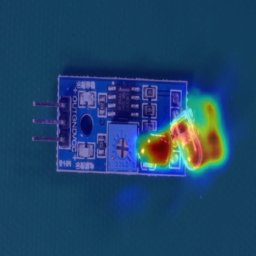}}\\

        {\includegraphics[width=1.\linewidth]{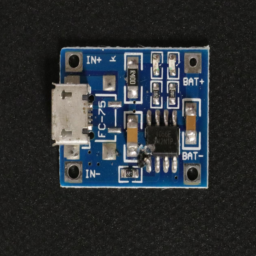}} &
        {\includegraphics[width=1.\linewidth]{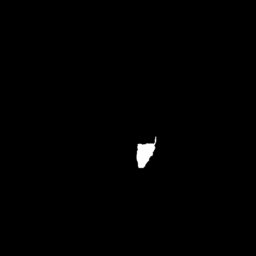}} &
        {\includegraphics[width=1.\linewidth]{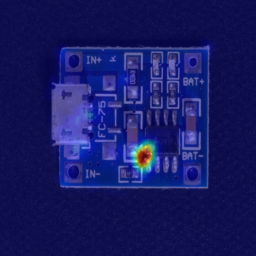}} &  
        \hspace{2.5cm}
        {\includegraphics[width=1.\linewidth]{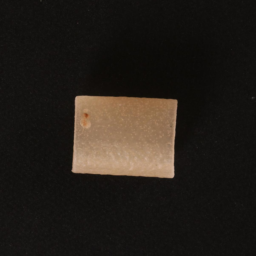}} &{\includegraphics[width=1.\linewidth]{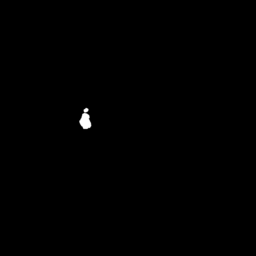}} &        {\includegraphics[width=1.\linewidth]{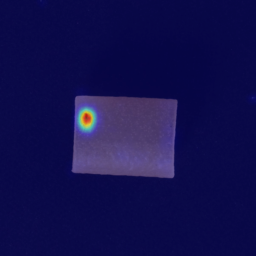}} &
        \hspace{2.5cm}
        {\includegraphics[width=1.\linewidth]{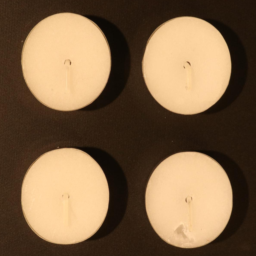}} &
        {\includegraphics[width=1.\linewidth]{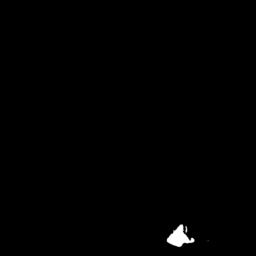}} &
        {\includegraphics[width=1.\linewidth]{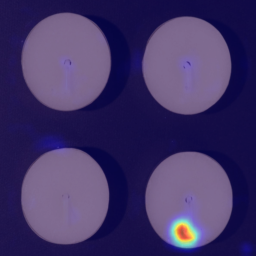}} &  
        \hspace{2.5cm}
        {\includegraphics[width=1.\linewidth]{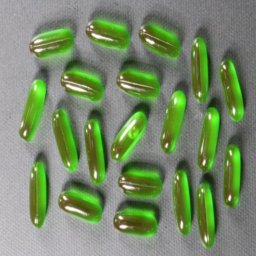}} &{\includegraphics[width=1.\linewidth]{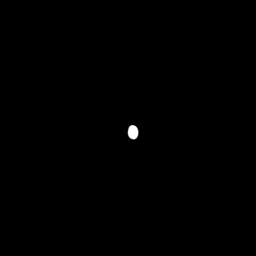}} &   {\includegraphics[width=1.\linewidth]{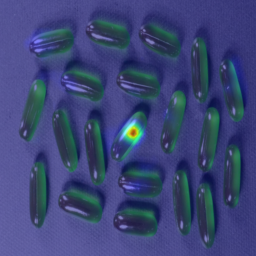}}\\
    \end{tabular}}
    \caption{The qualitative experimental results of anomaly maps. Columns 1, 4, 7 and 10 represent the original images, columns 2, 5, 8 and 11 show the ground truth masks, and columns 3, 6, 9 and 12 display the anomaly maps.}
    \label{fig:AblationC}
\end{figure*}

We displays qualitative reigon in ~\cref{fig:ablationA}, depicting the masks that represent the regions used in training when employing SCL. In addition, ~\cref{fig:AblationC} presents anomaly detection maps for the MVTec LOCO, MVTec AD, and VisA datasets.
% ------------------------------------------------------------------------------------

% ------------------------------------------------------------------------------------

\end{document}